%% file: main.tex
\documentclass{article}

     \usepackage[nonatbib,final]{main}

\usepackage{makecell}
\usepackage{graphicx}
\usepackage[utf8]{inputenc} 
\usepackage[T1]{fontenc}    
\usepackage{hyperref}       
\usepackage{url}            
\usepackage{booktabs}       
\usepackage{amsfonts}       
\usepackage{nicefrac}       
\usepackage{microtype}      
\usepackage{xcolor}
\usepackage{amsmath}
\usepackage{subfigure}
\usepackage{algorithm}
\usepackage{algorithmic}
\usepackage{amsmath,amssymb}
\usepackage{amsthm}
\usepackage{enumitem}
\usepackage{comment}
\usepackage{bm}
\usepackage{bbm}
\usepackage{xcolor}
\usepackage{makecell}
\usepackage[numbers]{natbib}

\newcommand{\Set}[1]{{\mathcal{#1}}}

\newcommand{\E}[0]{\mathbb{E}}
\renewcommand{\vec}[1]{{\bm{#1}}}

\definecolor{darkgreen}{rgb}{0,0.75,0}

\title{Is Independent Learning All You Need in the StarCraft Multi-Agent Challenge?}

\author{%
  Christian Schroeder de Witt\thanks{*: equal contribution. Corresponding author: Christian Schroeder de Witt <cs@robots.ox.ac.uk>. \newline \textsuperscript{$\dagger$} Department of Engineering Science, \textsuperscript{$\ddagger$} Department of Computer Science; University of Oxford, UK; \newline \textsuperscript{$\mathsection$} Snap Inc., USA; \textsuperscript{$\ddagger\ddagger$} NVIDIA, USA}$\mkern4mu$ \textsuperscript{$\dagger$} \\
  \And
  Tarun Gupta*\textsuperscript{$\ddagger$} \\
  \And
  Denys Makoviichuk\textsuperscript{$\mathsection$} \\
  \And
  Viktor Makoviychuk\textsuperscript{$\ddagger\ddagger$} \\
   \And
  Philip H.S. Torr\textsuperscript{$\dagger$} \\
   \And
  Mingfei Sun\textsuperscript{$\ddagger$} \\
   \And
  Shimon Whiteson\textsuperscript{$\ddagger$} \\
}

\begin{document}

\maketitle
\input{sections/00-abstract.tex}

\input{sections/01-introduction.tex}

\input{sections/02-related-work.tex}
\input{sections/03-preliminary.tex}
\input{sections/04-method.tex}
\input{sections/05-experiment.tex}
\input{sections/06-conclusion.tex}

\section{Acknowledgments and Disclosure of Funding}

We thank the members of the Whiteson Research Lab for their helpful feedback. This project hasreceived funding from the European Research Council (ERC), under the European Union’s Horizon2020 research and innovation programme (grant agreement number 637713). 
Tarun Gupta is supported by the Oxford University Clarendon Scholarship.  Mingfei Sun is supported by Microsoft Research Cambridge. The experiments were made possible by a generous equipmentgrant from NVIDIA. Philip Torr Shimon Whiteson is the Head of Research at Waymo, UK.

\newpage

\bibliographystyle{plain}
\bibliography{ref}

\newpage
\section{Appendix}
\label{sec:appendix}


\begin{figure}[h]
\begin{tabular}{ccc}
\label{fig:ippomore}
\includegraphics[width=45mm]{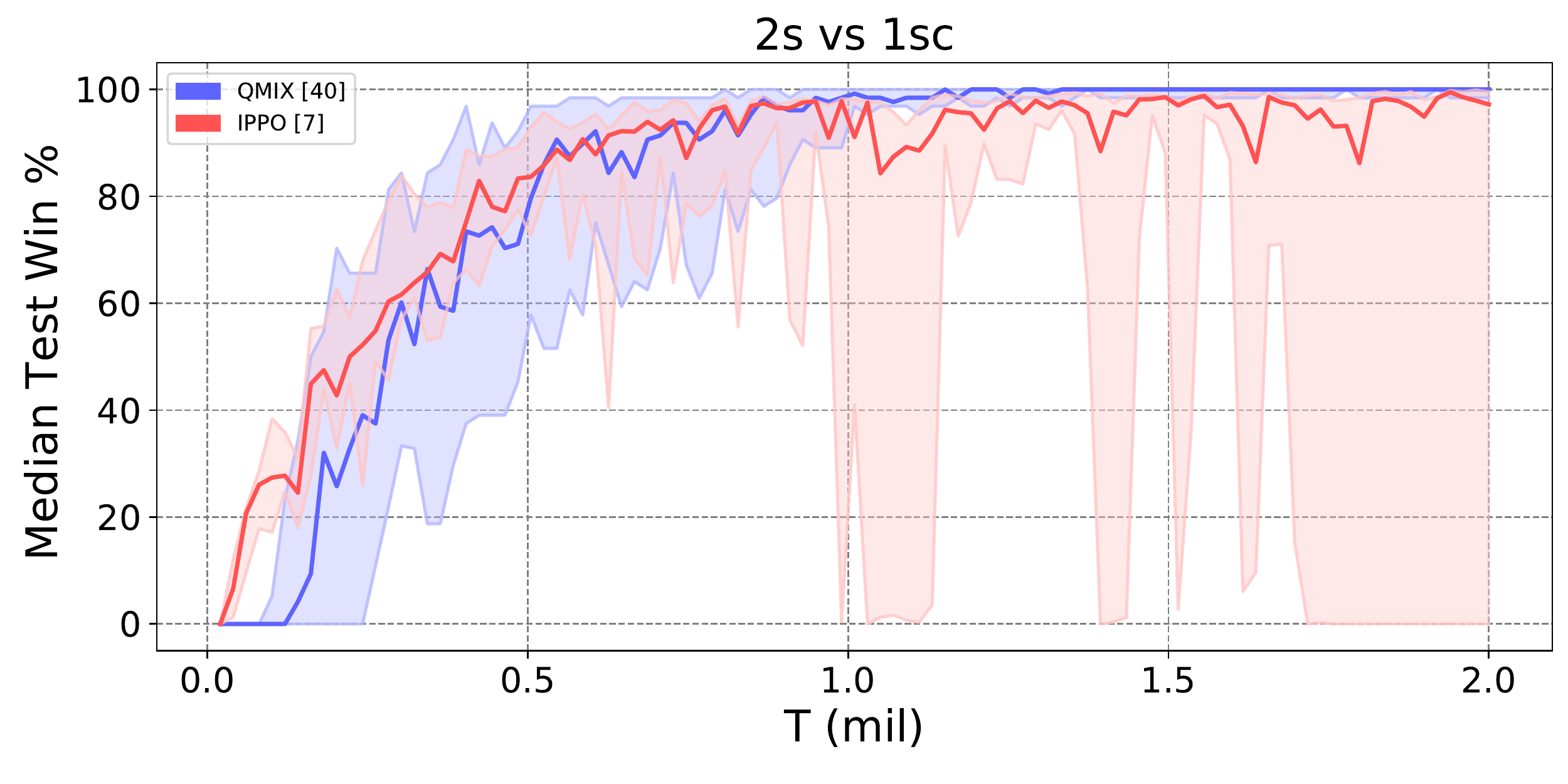} &
\includegraphics[width=45mm]{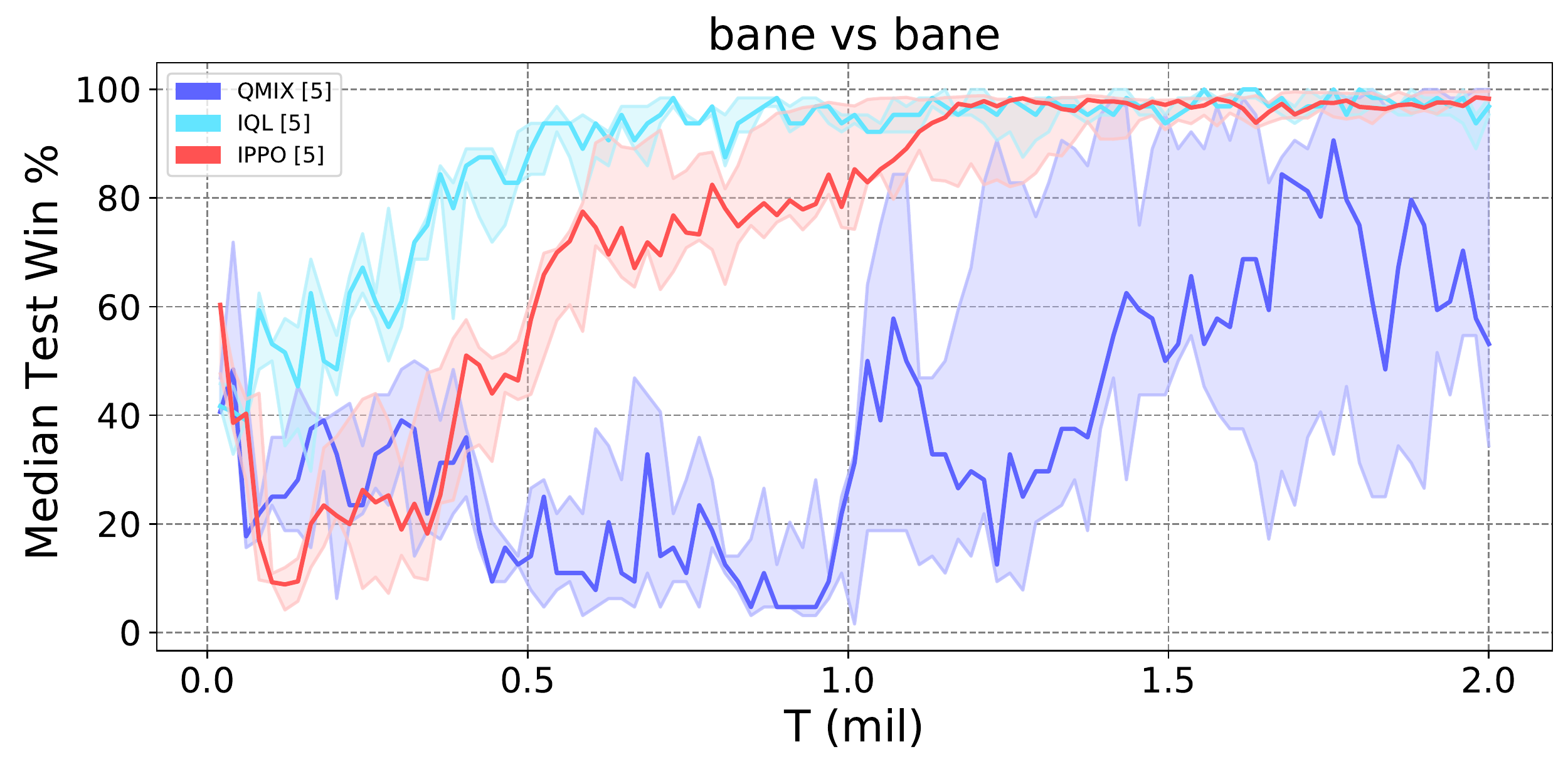} & 
\includegraphics[width=45mm]{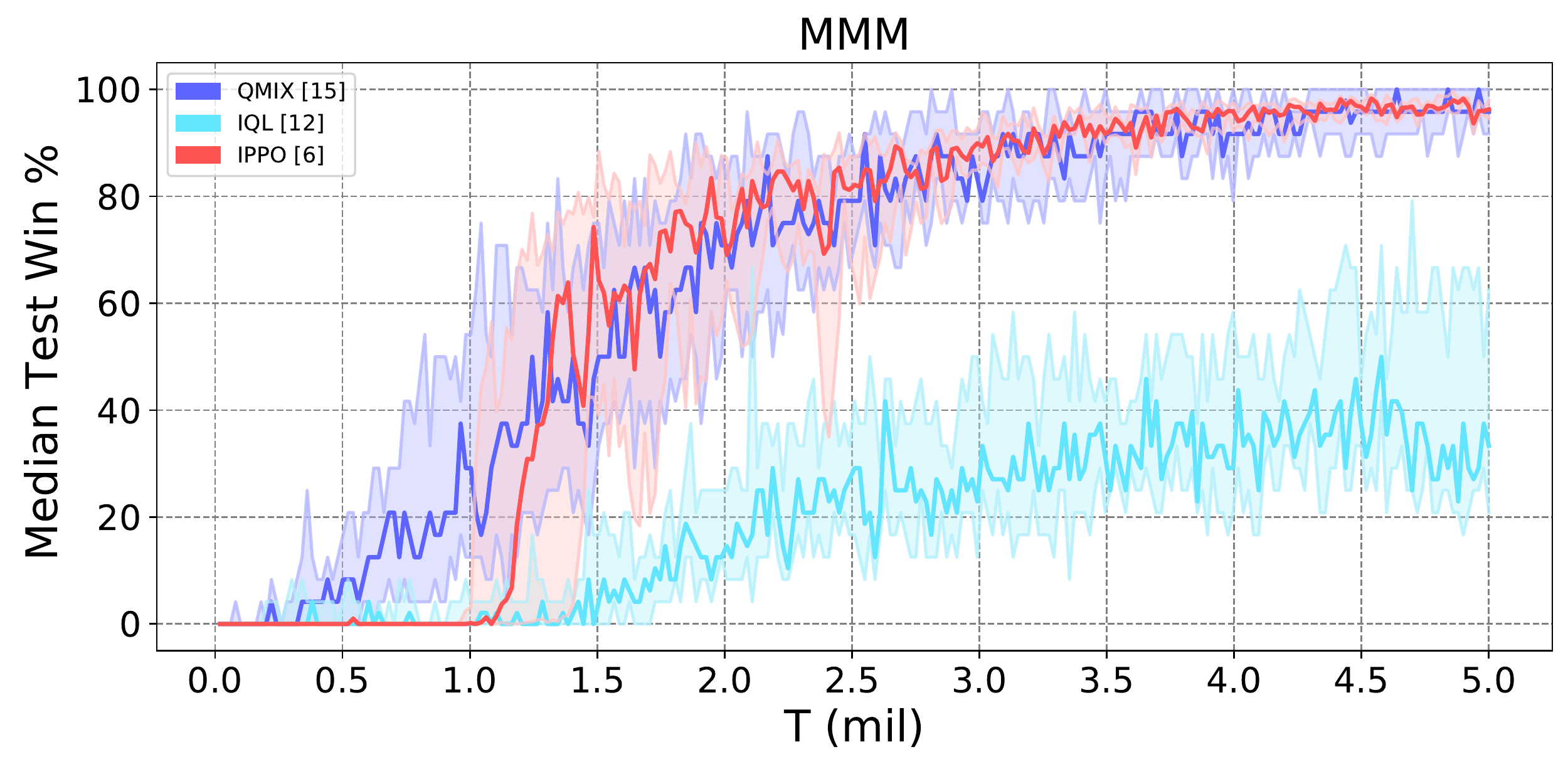}   \\
(a) 2s vs 1sc & (b) bane vs bane & (c) MMM \\[6pt]
\includegraphics[width=45mm]{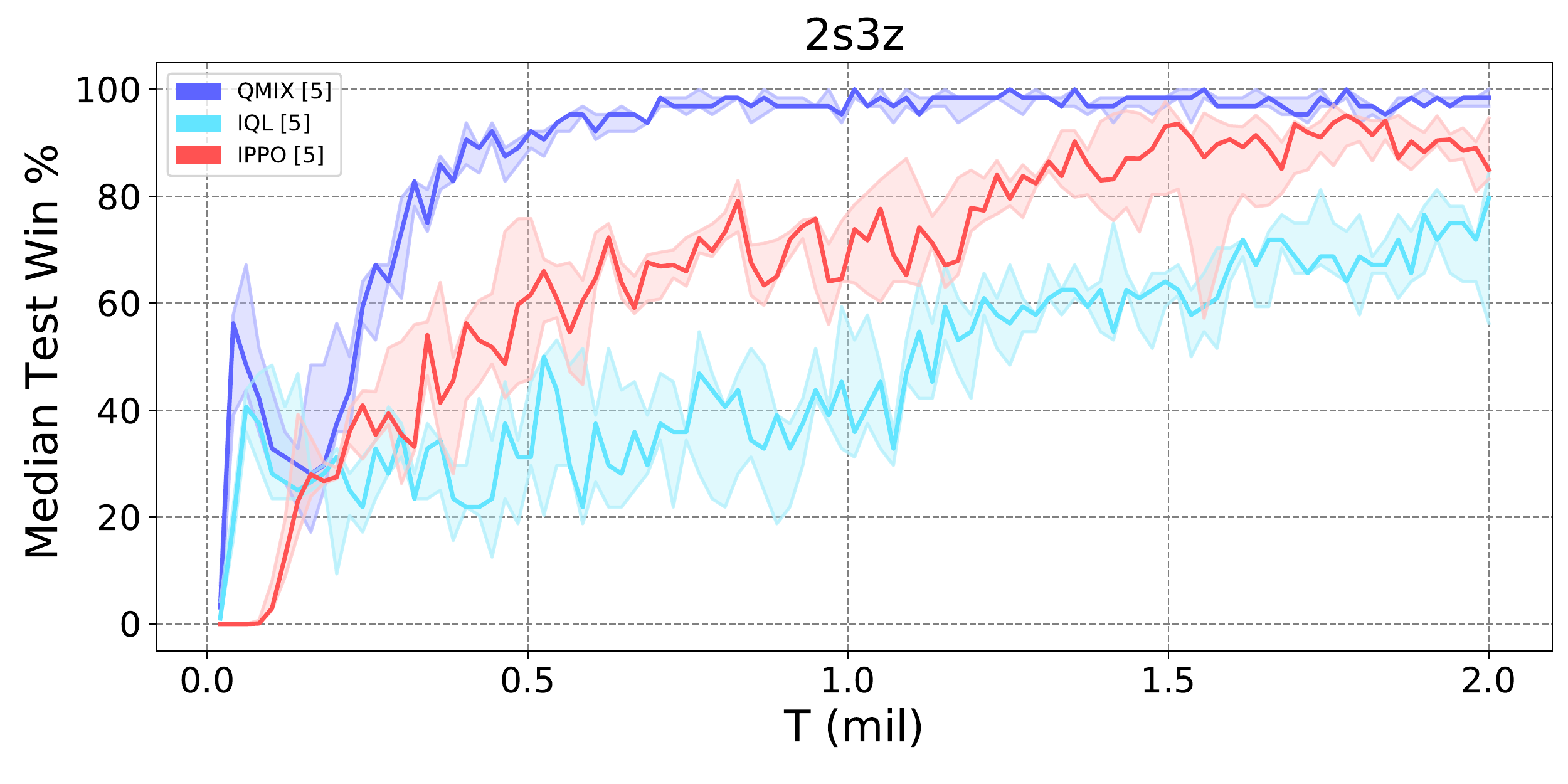}   & \includegraphics[width=45mm]{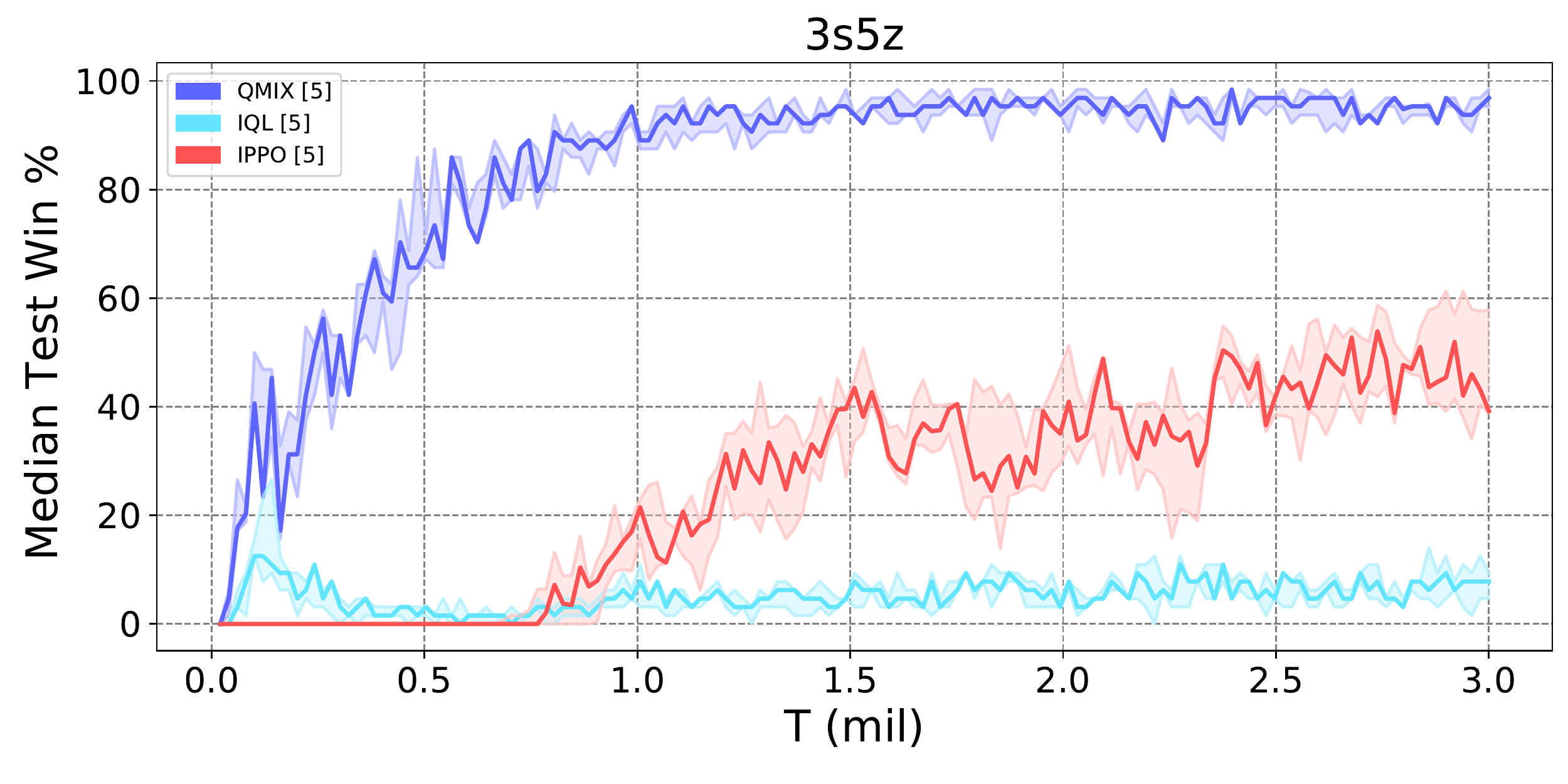} & \includegraphics[width=45mm]{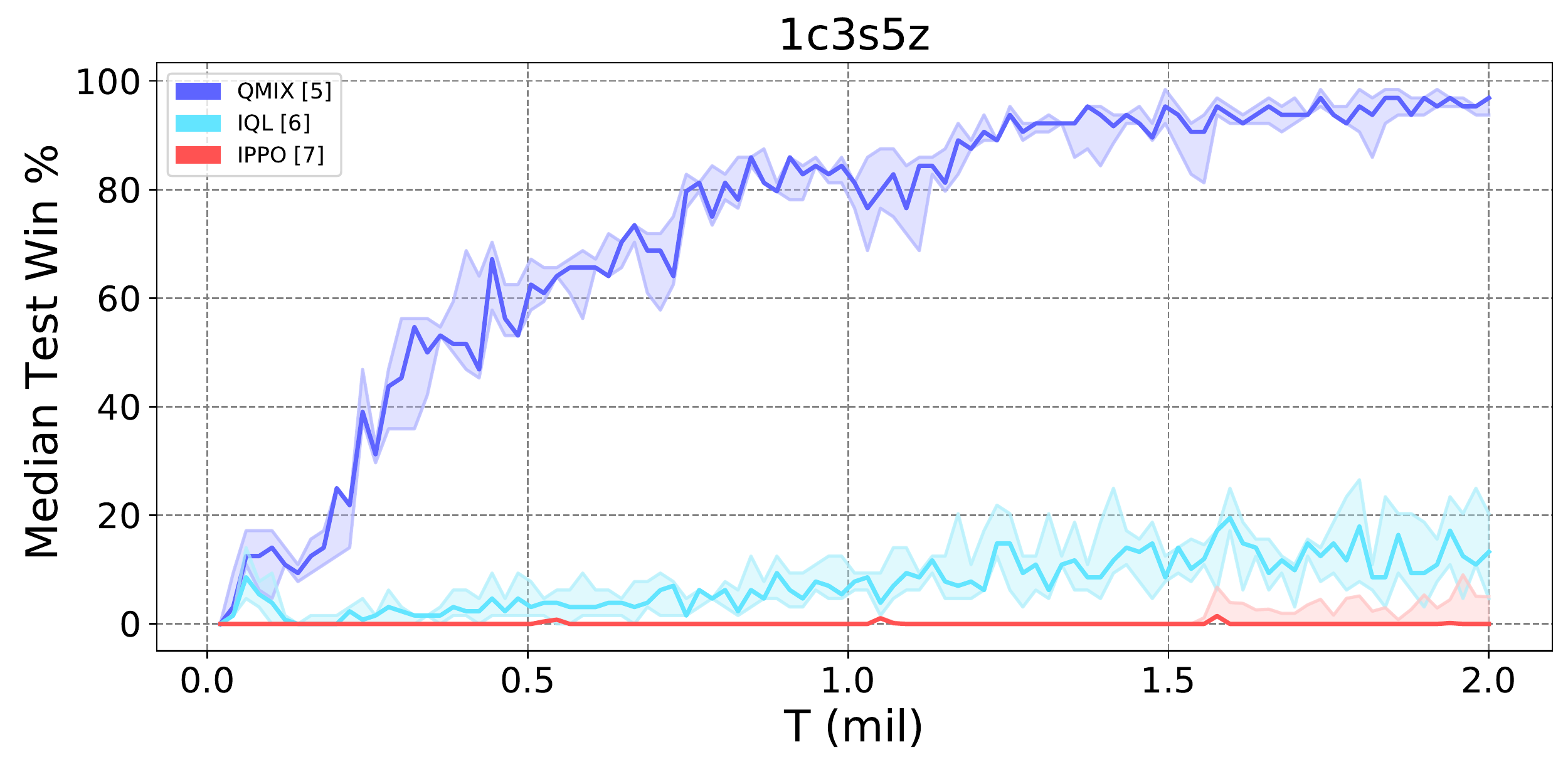} \\
(a) 2s3z & (b) 3s5z & (c) 1c3s5z \\[6pt]
\includegraphics[width=45mm]{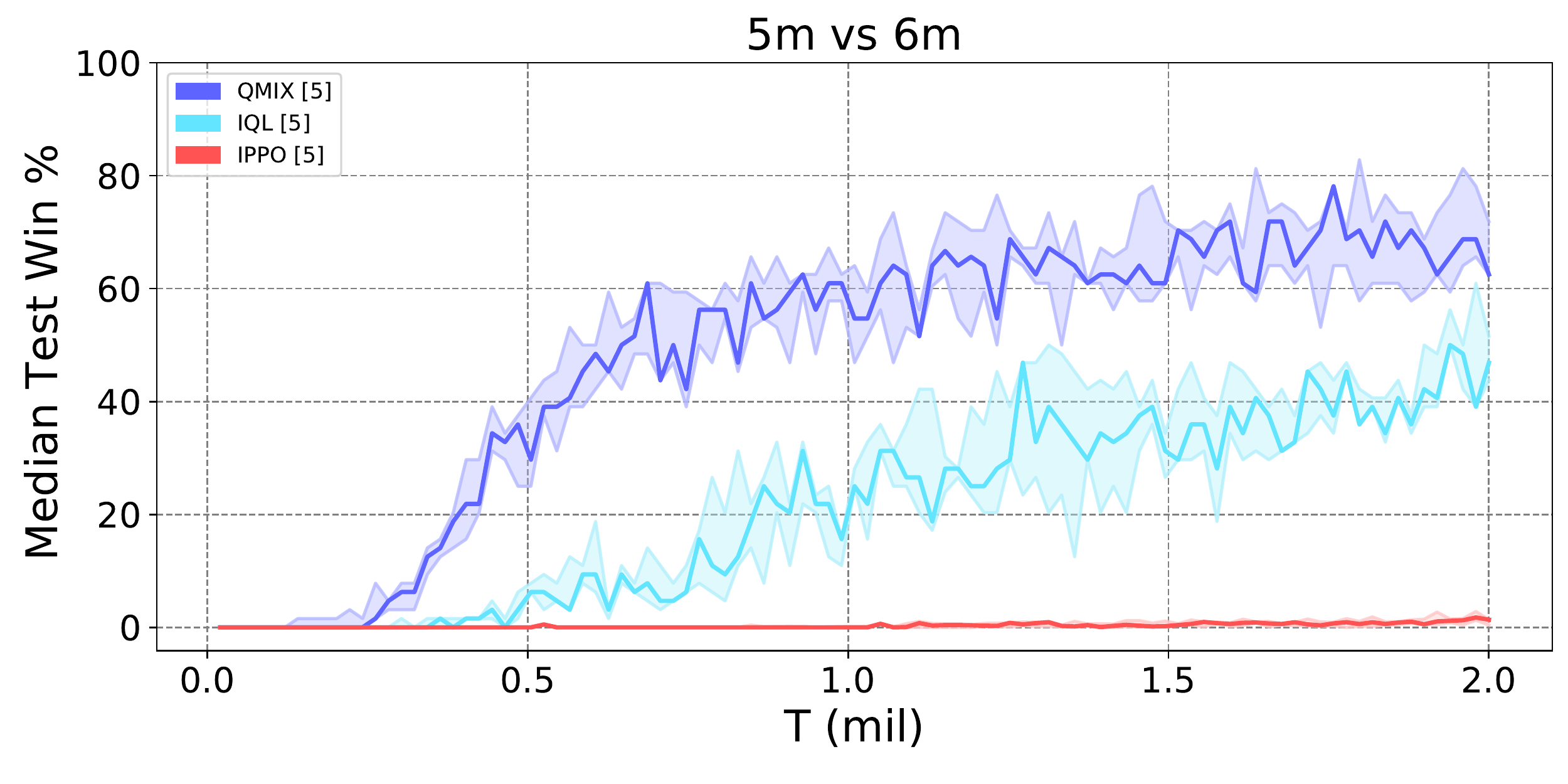} &
\includegraphics[width=45mm]{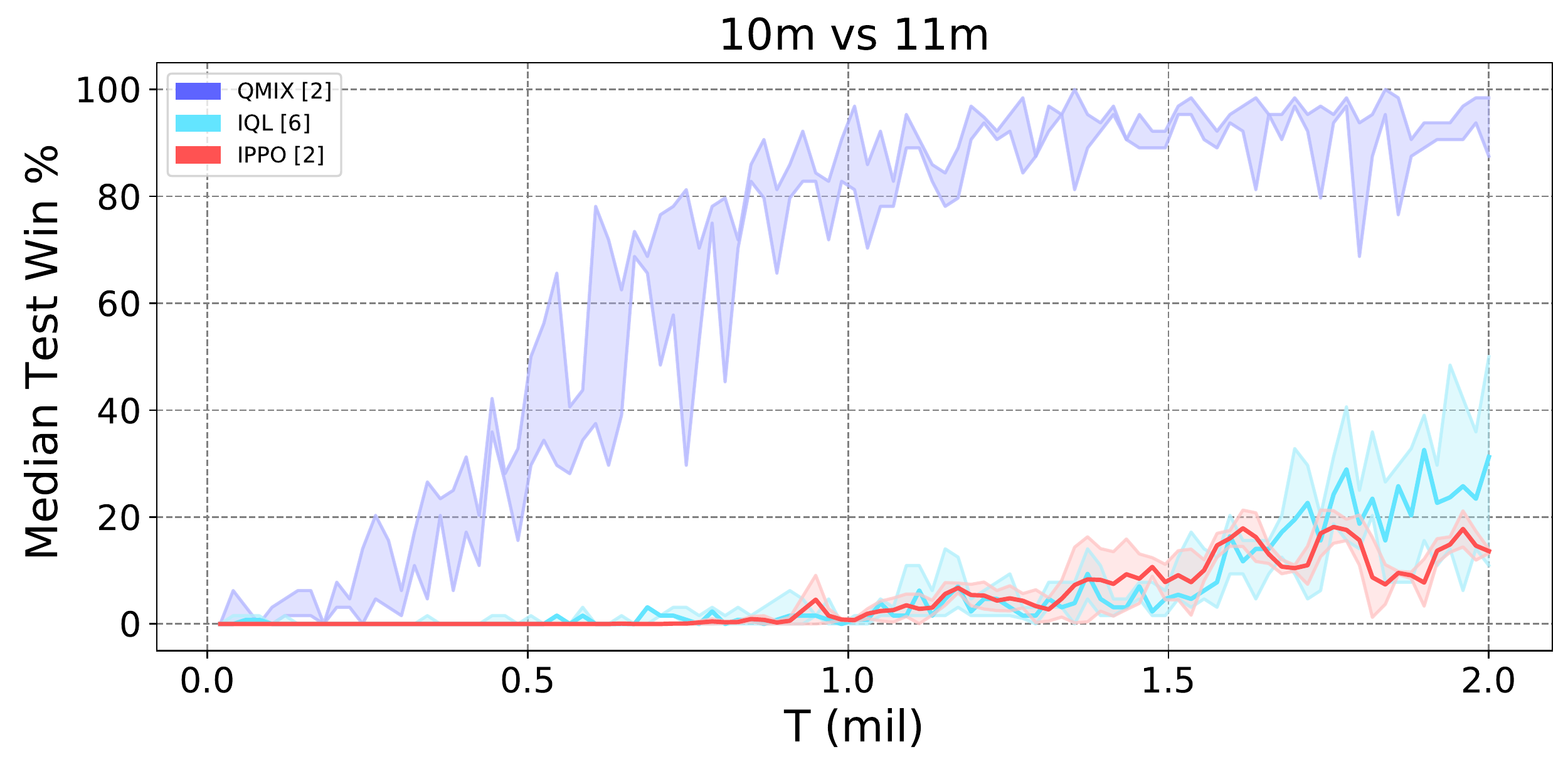} & \includegraphics[width=45mm]{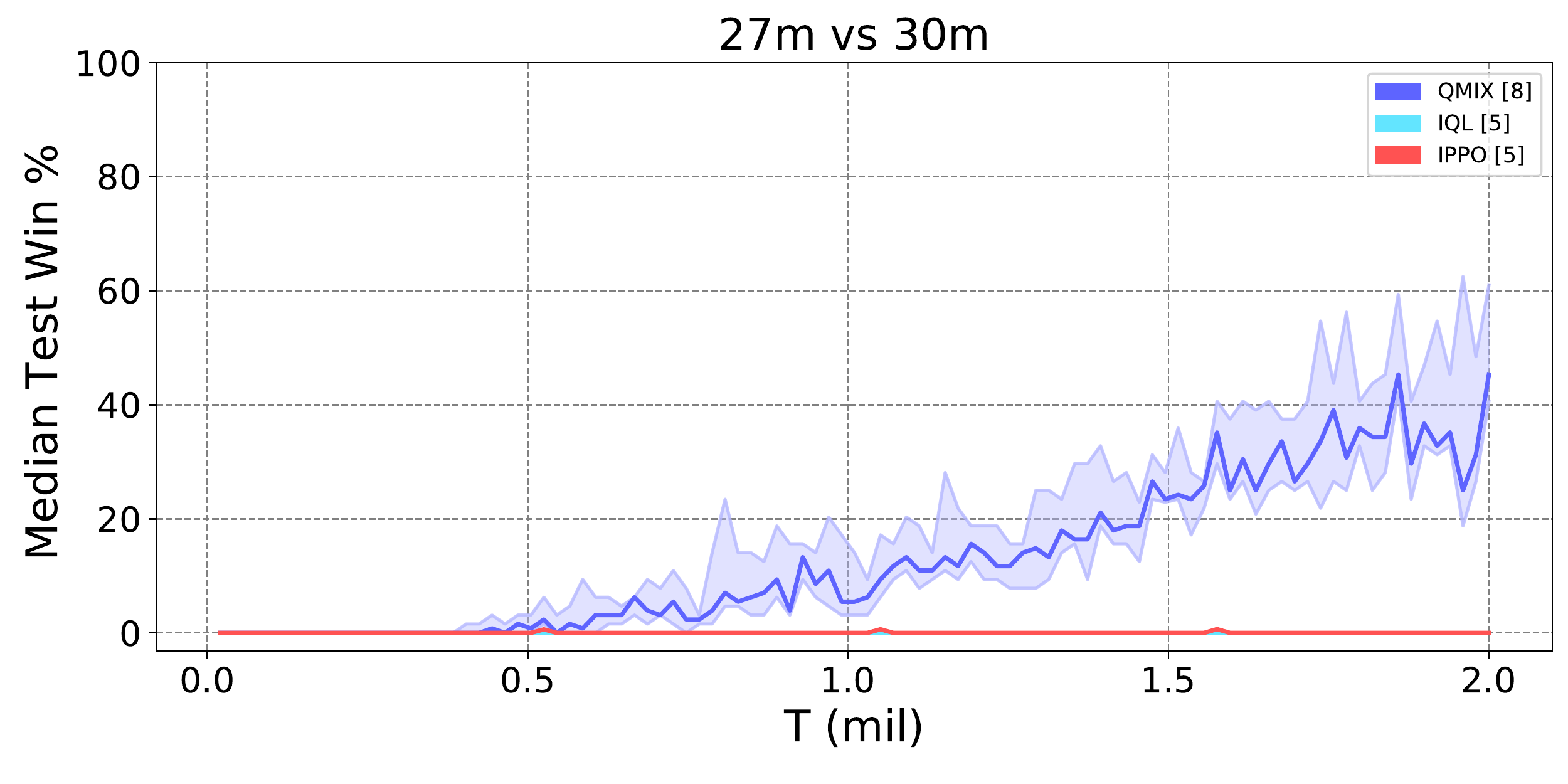}
 \\
(g) 5m vs 6m & (h) 10m vs 11m & (i) 27m vs 30m  \\[6pt]
\multicolumn{3}{c}{
\begin{tabular}{cc}
\includegraphics[width=45mm]{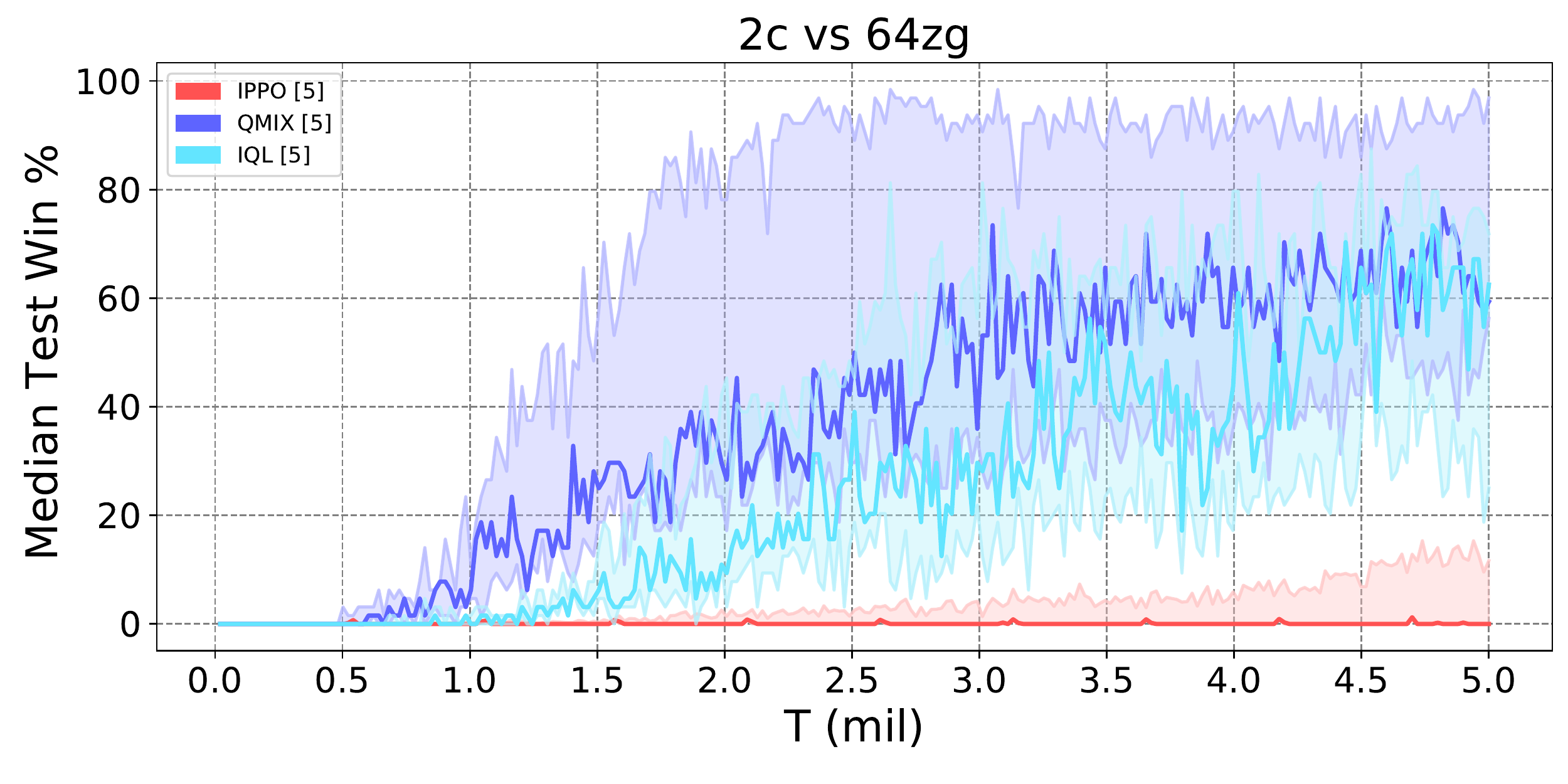} & \includegraphics[width=45mm]{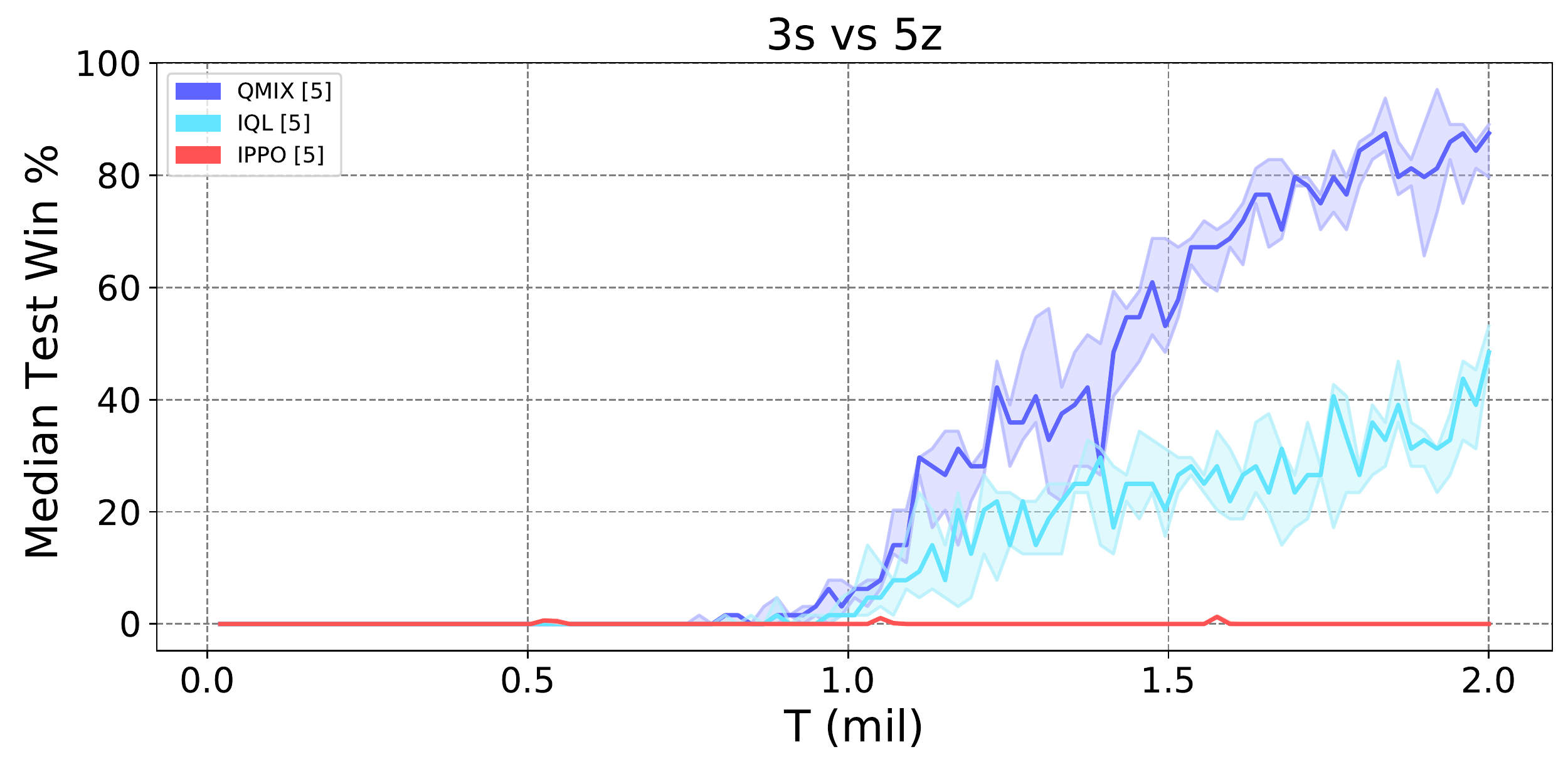} \\
2c vs 64zg &  3s vs 5z \\
\end{tabular}
}

\end{tabular}
\caption{Results on select SMAC maps comparing IPPO and IPPO-C with QMIX, VDN and IQL}
\end{figure}

\begin{figure}[h]
\label{fig:over}
\begin{center}
 \includegraphics[width=65mm]{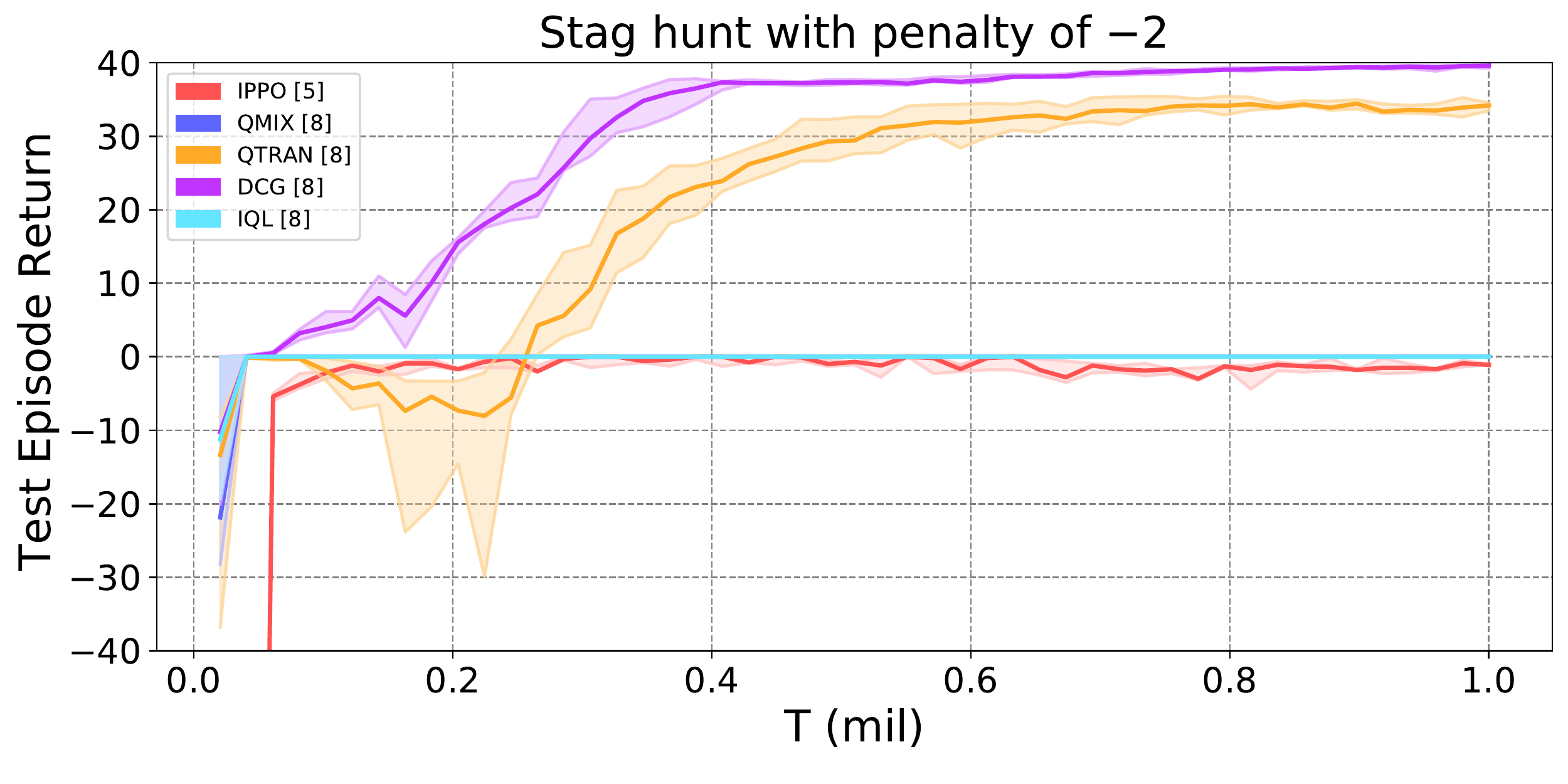} \\ 
\end{center}
\caption{Relative overgeneralization, as experienced by select MARL algorithms on Stag Hunt with penalty $-2$ (see \cite{bohmer_deep_2020}). Note that DCG is not fully decentralized during execution. }
\end{figure}

\end{document}

%% file: sections/00-abstract.tex
\begin{abstract}
   Most recently developed approaches to cooperative multi-agent reinforcement learning in the \emph{centralized training with decentralized execution} setting involve estimating a centralized, joint value function.  In this paper, we demonstrate that, despite its various theoretical shortcomings, Independent PPO (IPPO), a form of independent learning in which each agent simply estimates its local value function, can perform just as well as or better than state-of-the-art joint learning approaches on popular multi-agent benchmark suite SMAC with little hyperparameter tuning. We also compare IPPO to several variants; the results suggest that IPPO's strong performance may be due to its robustness to some forms of environment non-stationarity.
\end{abstract}

%% file: sections/01-introduction.tex
\section{Introduction}

 Many practical control problems feature a team of multiple agents that must coordinate to achieve a common goal \cite{cao_overview_2012,huttenrauch_guided_2017}. 
 Cooperative multi-agent reinforcement learning (MARL) has shown considerable promise in solving tasks that can be described as a Dec-POMDP \cite{oliehoek_concise_2016}, 
 i.e., where agents optimise a single scalar team reward signal in a partially observable environment while choosing actions based only on their own local action-observation histories \cite{panait_cooperative_2005,busoniu_comprehensive_2008,tuyls_multiagent_2012}. 

Independent learning (IL) decomposes an $n$-agent MARL problem into $n$ decentralised single-agent problems where all other agents are treated as part of the environment, and learning policies that condition only on an agent's local observation history. 
 While easy to distribute and decentralisable by construction, IL suffers from a variety of theoretical limitations that may result in learning instabilities or suboptimal performance observed in practice \cite[IQL, IAC]{tan_multi-agent_1993,foerster_counterfactual_2017}. 
 Firstly, the presence of other learning and exploring agents renders the resulting environment non-stationary from the given agent's perspective, forfeiting convergence guarantees \cite{tan_multi-agent_1993}. 
 Secondly, independent learners are not always able to distinguish environment stochasticity from another agent's exploration, making them unable to learn optimal policies in some environments \cite{claus_dynamics_1998}. 

In fact, decentralised policies need not be learnt in a decentralized fashion. 
 For safety and efficiency reasons \cite{sutton_reinforcement_2018}, MARL training frequently takes place centrally in a laboratory or in simulation, 
 allowing agents access to each other's observations during training, as well as otherwise unobservable extra state information. 
 Centralized training allows training of a single joint policy for all agents that conditions on the joint observations and extra state information. 
 While centralized joint learning reduces or removes issues surrounding partial observability and environment non-stationarity, 
 it must cope with joint action spaces that grow exponentially with respect to the number of agents, as well as a variety of learning pathologies that can result in suboptimal policies \cite{wei_lenient_2016}. 
 Importantly, vanilla joint policies are not inherently decentralisable and naive policy distillation approaches are often ineffective \cite{bohmer_exploration_2019}. 
Joint learning does not immediately address the multi-agent credit assignment problem either.

Recent research has focused on algorithms that can exploit the benefits of combining centralised training with decentralised execution \cite[CTDE]{oliehoek_concise_2016}, 
 while mitigating some of the pathologies of vanilla centralized joint learning. A particularly successful line of research uses value factorisation of the joint $Q$-value function in order to reduce the size of the joint action space [VDN, QMIX, FacMADDPG]\cite{sunehag_value-decomposition_2017, rashid_monotonic_2020, de_witt_deep_2020}.

However, value factorization is prone to a learning pathology called \textit{relative overgeneralization} \cite{wei_lenient_2016} where policies erroeneously converge to a suboptimal joint action.
Relative overgeneralisation commonly arises when multiple agents must coordinate their actions but receive negative rewards if only a subset of them do so. 
 In this case, random occurrences of such successful coordination events are a needle in a haystack and monotonic value factorisations cannot represent the non-monotonic team reward function \cite{bohmer_deep_2020}. 
 A variety of recent approaches employ joint optimisation of, or transfer learning between, factored and unrestricted joint value functions in order to overcome the representational limitations of factored value functions, with varying success \cite{bohmer_exploration_2019,son_qtran_2019}. 

In this paper, we present empirical evidence for Independent PPO (IPPO), a multi-agent variant of \textit{proximal policy optimization} \cite{schulman2017proximal}, 
 that shows IPPO matches or outperforms state-of-the-art MARL CTDE algorithms such as QMIX \cite{rashid_monotonic_2020} or MAVEN \cite{mahajan_maven_2020} on multiple hard maps on SMAC \cite{samvelyan_starcraft_2019}.


Given the purported pathologies of existing IL approaches such as IQL and IAC, we hypothesise that algorithmic choices made by PPO such as policy clipping help mitigate some forms of environment non-stationarity, rendering fundamental theoretical limitations less important to practical performance. We empirically show that the effect of policy clipping in IPPO cannot be emulated by lowering the effective learning rate alone.

%% file: sections/02-related-work.tex
\section{Related Work}

 Actor-critic algorithms have been shown to benefit from learning centralized joint critics alongside decentralised policies \cite[Central-V]{foerster_counterfactual_2017}. 
 As the critics are not needed during execution, this approach is inherently decentralizable. 
COMA \cite{foerster_counterfactual_2017} extends this approach with a counterfactual multi-agent critic baseline based on temporal difference errors \cite{tumer_distributed_2007} in order to facilitate multi-agent credit assignment.  
Joint $Q$-learning can also be made decentralizable. Value Decomposition Networks \cite[VDN]{sunehag_value-decomposition_2017} decompose joint state-action value functions into sums of decentralised utility functions that can be used during greedy execution. 
 QMIX \cite{rashid_monotonic_2020}  extends this additive decomposition to arbitrary centralized monotonic mixing networks. 
 Both centralized joint critics and factored joint value functions can reap some benefits of centralized joint learning, 
 while bypassing the joint action space explosion, imposing an effective prior on multi-agent credit assignment, and mitigating practical learning pathologies associated with centralized joint learning on popular benchmark environment StarCraft II \cite[SMAC]{samvelyan_starcraft_2019}. 
Value factorisation has also been successfuly transferred to continuous action spaces and combined with actor critic approaches \cite[COMIX, FacMADDPG]{de_witt_deep_2020}.

Independent learning (IL) dates back to the early days of multi-agent reinforcement learning \cite[IQL]{tan_multi-agent_1993}. While initially tabular, IL algorithms using neural networks as function approximators were subsequently developed [IAC, deep IQL]\cite{rashid_monotonic_2020, foerster_counterfactual_2017}.
The question of whether independent agents are better at learning cooperative behaviour than multiple \textit{complete observing} agents \cite{whitehead_complexity_1991} was first comprehensively investigated by Tan \cite{tan_multi-agent_1993}, who concludes that while sharing policies or experience was generally advantageous, sharing extra sensory information could also negatively interfere with learning, e.g., by expanding the state space. 

Trust region optimisation for reinforcement learning was popularised by TRPO \cite{schulman_trust_2017} which implements iterative guaranteed monotonic improvements. PPO \cite{schulman2017proximal} preserves many of TRPO's empirical benefits while trading in theoretical guarantees for computational speed. PPO's policy update regularisation using clipped probability ratios has been subject to much scrutiny in single-agent settings: \textit{Truly PPO} \cite{wang_truly_2020} suggests modifications in order to ensure guaranteed monotonic improvements with little computational overhead. Recent work confirms that PPO's performance may crucially depend on the choice of policy surrogate objective \cite{hsu_revisiting_2020}. In addition, code-level design choices have been shown to significantly impact PPO performance in practice \cite{engstrom_implementation_2020}.

While popular for single agent tasks, PPO has only recently been applied to decentralised cooperative multi-agent tasks. Concurrent work proposes MAPPO \cite{anonymous_benchmarking_2020}, an actor-critic multi-agent algorithm based on PPO. Like IPPO, MAPPO employs weight sharing between each agent's critic. In contrast to IPPO, MAPPO uses a centralized value function that conditions on the full  state (or concatenation of agent observations, if the full state is unavailable). As such, MAPPO is not an \textit{independent} learning algorithm. We show that IPPO substantially outperforms MAPPO on a variety of hard SMAC maps and even beats other state-of-the-art algorithms, such as QMIX, on some of these, with only modest hyperparameter tuning. 



%% file: sections/03-preliminary.tex

\section{Background}
\label{sec:background}

\paragraph{Dec-POMDPs.} We consider a {\em fully cooperative multi-agent task} 
A {\em decentralised partially observable Markov decision process} 
\citep[Dec-POMDP][]{oliehoek_concise_2016} describes multi-agent tasks where  a team of cooperative agents chooses sequential actions under partial observability and environment stochasticity.
Dec-POMDPs can be formally defined by a tuple $\langle \mathcal{N} , \Set S, \Set U, P, r, \Set Z, O, \rho, \gamma \rangle$. 
Here $s \in \Set S$ describes the state of the environment, discrete or continuous, and $ \mathcal{N} := \{1,\dots,N\}$ denotes the set of $N$ agents. $s_0 \sim \rho$, the initial state, is drawn from distribution $\rho$. At each time step $t$, all agents $a \in \mathcal{N}$ simultaneously 
choose actions $u^a_t \in \Set U$ which may be discrete or continuous. This yields the joint action $\vec{u}_t := \{u_t^a\}_{a=1}^{N} \in \Set U^N$. 
The next state $s_{t+1} \sim P(s_t, \vec u_t)$ is drawn from transition kernel $P$ after executing the joint action $\vec u_t$ in state $s_t$. Subsequently, the agents receive a scalar team reward $r_t = r(s_t, \vec u_t)$.

Instead of being able to observe the full state $s_t$, in a Dec-POMDP each agent $a \in \mathcal{N}$ can only draws an individual 
local observation $z_t^a \in \Set Z, \vec z_t := \{z_t^a\}_{a=1}^N$,  
from the observation kernel $O(s_t, a)$. The history of an agent's observations and actions is denoted by
$\tau^a_t \in \Set T_t := (\Set Z \times \Set U)^t \times \Set Z$.
The set of all agents' histories is given by $\vec\tau_t := \{\tau^a_t\}_{a=1}^N$.
Each agent $a$ chooses its actions with a decentralised 
policy $u^a_t \sim \pi^a(\cdot|\tau^a_t)$ that is
based only on its individual history. 

The team of cooperative agents attempts to learn 
a {\em joint policy} $\pi(\vec u|\vec\tau_t) 
:= \prod_{a=1}^N \pi^a(u^a|\tau^a_t)$
that maximises their expected discounted return, 
$J(\vec \pi) \doteq \E[\sum_{t=0}^{\infty} \gamma^t r_t]$, where $\gamma \in [0, 1)$ is a discount factor. 
$\pi(\vec u|\vec\tau_t)$ induces a joint action-value function $Q^\pi$
that estimates the expected discounted return of joint action $\vec{u}_t$ 
in state $s_t$ with histories $\vec \tau_t$. Using the joint action-value function, agents then follow a joint policy $\pi$ using $Q^\pi(s_t,\vec\tau_t, \vec{u}_t) :=\mathbb{E}\left[\sum_{i=0}^{\infty}\gamma^ir_{t+i}\right]$.

\paragraph{Centralised learning with decentralised execution.} Reinforcement learning policies can often be learnt in simulation or in a laboratory. In this case, on top of their local observation histories, agents may have access to the full environment state and share each other's policies and experiences during training. The framework of \textit{centralised training with decentralised execution (CTDE)} \cite{oliehoek_concise_2016,kraemer_multi-agent_2016} formalises the use of centralised information to facilitate the training of decentralisable policies.

\paragraph{Trust Region Policy Optimization} TRPO \citep{schulman_trust_2017} is a class of policy-gradient methods that restricts the update of policies to within the trust region of the behavior policy by enforcing a KL divergence constraint on policy update at each iteration. Formally, TRPO optimizes the following:
\begin{align}
\label{eq:trpo}
&\max_{\theta} \qquad \qquad \mathbb{E}_{s_t, u_t} \left[ \frac{\pi_{\theta}(u_t | s_t)}{\pi_{\theta_{old}}(u_t | s_t)} A(s_t, u_t) \right] \\
&\textrm{subject to} \qquad\,\, \mathbb{E}_{s_t, u_t}\left[\textrm{KL}\left(\pi_{\theta_{old}}, \pi_{\theta}\right)\right] \leq \delta,
\end{align}
where $\theta_{old}$ are the policy parameters before the update and $A(s_t, u_t)$ is an approximation of advantage function. 
The formulation is computationally expensive as it requires the computation of multiple Hessian-vector products for nonlinear conjugate gradients when approximating the KL constraint. 
To resolve this, Proximal Policy Optimization \citep{schulman2017proximal} approximates the trust region constraints by policy ratio clippings, i.e. the policy loss becomes:
\begin{align}
\label{eq:ippo}
\mathcal{L}(\theta) = \mathbb{E}_{s_t, u_t} \left[ \min\left( \frac{\pi_{\theta}(u_t | s_t)}{\pi_{\theta_{old}}(u_t | s_t)} A(s_t, u_t), \textrm{clip}\left(\frac{\pi_{\theta}(u_t | s_t)}{\pi_{\theta_{old}}(u_t | s_t)}, 1-\epsilon, 1+\epsilon\right) A(s_t, u_t) \right) \right].
\end{align}

\raggedbottom

%% file: sections/04-method.tex
\section{Independent PPO}


In this paper, we use PPO to learn decentralized policies $\pi^a$ for agents with individual policy clipping where each agent's independent policy updates are clipped based on the objective defined in Equation \ref{eq:ippo}. We consider a variant of the advantage function based on independent learning, where each agent $a$ learns a local observation based critic $V_{\phi}(z_t^a)$ parameterised by $\phi$ using \textit{Generalized Advantage Estimation} (GAE) \citep{schulman_high-dimensional_2018} with discount factor $\gamma=0.99$ and $\lambda=0.95$. 
The network parameters $\phi, \theta$ are shared across critics, and actors, respectively. 
We also add an entropy regularization term to the final policy loss \citep{mnih_asynchronous_2016}.  
For each agent $a$, we have its advantage estimation as follows:
\begin{align}
A^{a}_t = \sum_{l=0}^{h} (\gamma \lambda)^l \delta_{t+l}^{a},
\end{align}
where $\delta_{t}^{a} = r_t(z_{t}^a, u_{t}^a) + \gamma V_{\phi}(z_{t+1}^a) - V_{\phi}(z_t^a)$ is the TD error at time step $t$ and we vary $h$ as shown in Table \ref{table:paramtable} (marked as steps num).
We use the team reward $r_t(s_t, a_t)$ to approximate $r_t(z_{t}^a, u_{t}^a)$. The final policy loss for each agent $a$ becomes:
\begin{align}
\mathcal{L}^a(\theta) = \mathbb{E}_{z_t^a, u_t^a} \left[ \min\left( \frac{\pi_{\theta}(u_t^a | z_t^a)}{\pi_{\theta_{old}}(u_t^a | z_t^a)} A_t^a, \,\, \textrm{clip}\left(\frac{\pi_{\theta}(u_t^a | z_t^a)}{\pi_{\theta_{old}}(u_t^a | z_t^a)}, 1-\epsilon, 1+\epsilon\right) A_t^a \right) \right].
\end{align}

\textbf{Value Clipping}: In addition to clipping the policy updates, our method also uses value clipping to restrict the update of critic function for each agent $a$ to be smaller than $\epsilon$ as proposed by \cite[GAE]{schulman_high-dimensional_2018} using:
\begin{align}
\mathcal{L}^a(\phi) = \mathbb{E}_{z_t^a} \left[ \min \left\lbrace \left(V_{\phi}(z_t^a) - \hat{V}_t^a \right)^2,  \left(V_{\phi_{old}}(z_t^a) + \textrm{clip}(V_{\phi}(z_t^a) - V_{\phi_{old}}(z_t^a), -\epsilon, +\epsilon) - \hat{V}_t^a \right)^2\right\rbrace \right],
\end{align}
where $\phi_{old}$ are old parameters before the update and $\hat{V}_t^a = A_t^a + V_{\phi}(z_t^a)$. 
The update equation restricts the update of the value function to within the trust region, and therefore helps us to avoid overfitting to the most recent batch of data. 
For each agent, the overall learning loss becomes:
\begin{align}
 \mathcal{L}(\theta, \phi) = \sum_{a=1}^n \mathcal{L}^a(\theta) + \lambda_{\textrm{critic}} \mathcal{L}^a(\phi) + \lambda_{\textrm{entropy}} \mathcal{H}(\pi^a),
 \end{align}
  where $\mathcal{H}(\pi^a)$ denotes the entropy of policy $\pi^a$, $\lambda_{\textrm{critic}}$ and $\lambda_{\textrm{entropy}}$ vary as shown in Table \ref{table:paramtable}. In Section \ref{sec:ablation}, we compare IPPO to a baseline in which both policy and value clipping are ablated, yielding a variant of IAC  \cite{foerster_counterfactual_2017}. 


\textbf{Learning Architecture}: We use a variance scaling initializer with truncated normal distribution \cite{he2015delving} with $\textrm{scale}=2.0$ to initialize the parameters of our policy and value function NNs that has been shown to work well with ReLU activations. The input to our NN comprises of stacked observations for the past few times steps (marked as frames in Table \ref{table:paramtable}), which is passed through three conv1d convolution layers with varying number of filters (marked as net arch in Table) and fixed kernel\_size = 3, strides = (2, 1, 1), padding = (same, valid, valid) for each conv layer, followed by two MLP layers with (256, 128) units and ReLU activations (see Table \ref{table:paramtable} for more details). We use gradient clipping to restrict the norm of the gradient to be less than $0.5$ and normalize the advantage by subtracting the mean and dividing by the standard deviation once before training - we find that advantage normalisation by minibatch yields worse performance. We use a discount factor of $\gamma=0.99$, a learning rate of $10^{-4}$, and clipping parameter $\epsilon=0.2$.

%% file: sections/05-experiment.tex
\section{Empirical Results}
\label{sec:exps}

\begin{figure}
\label{fig:ippo}
\begin{tabular}{cc}
  \includegraphics[width=65mm]{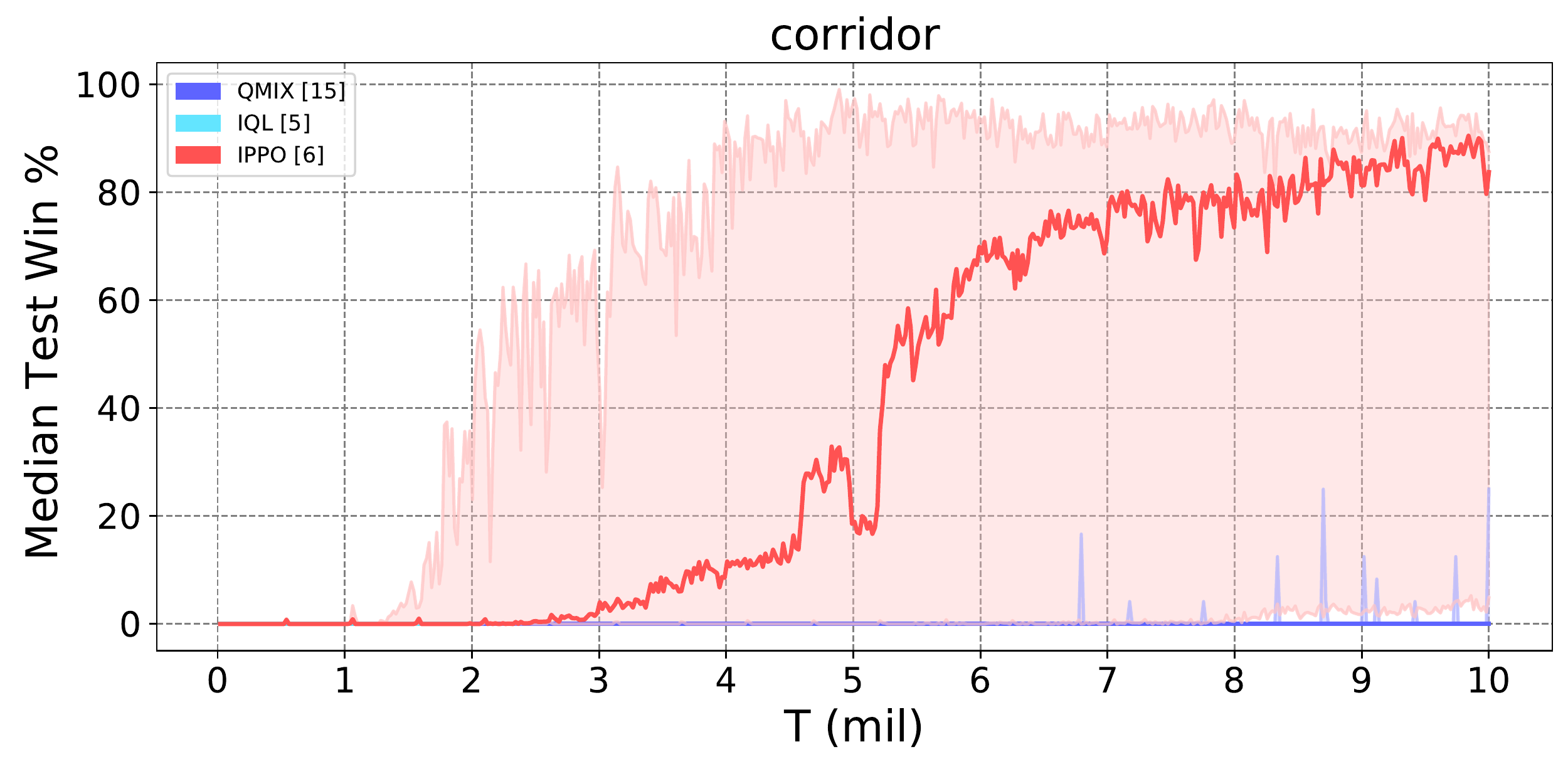} & \includegraphics[width=65mm]{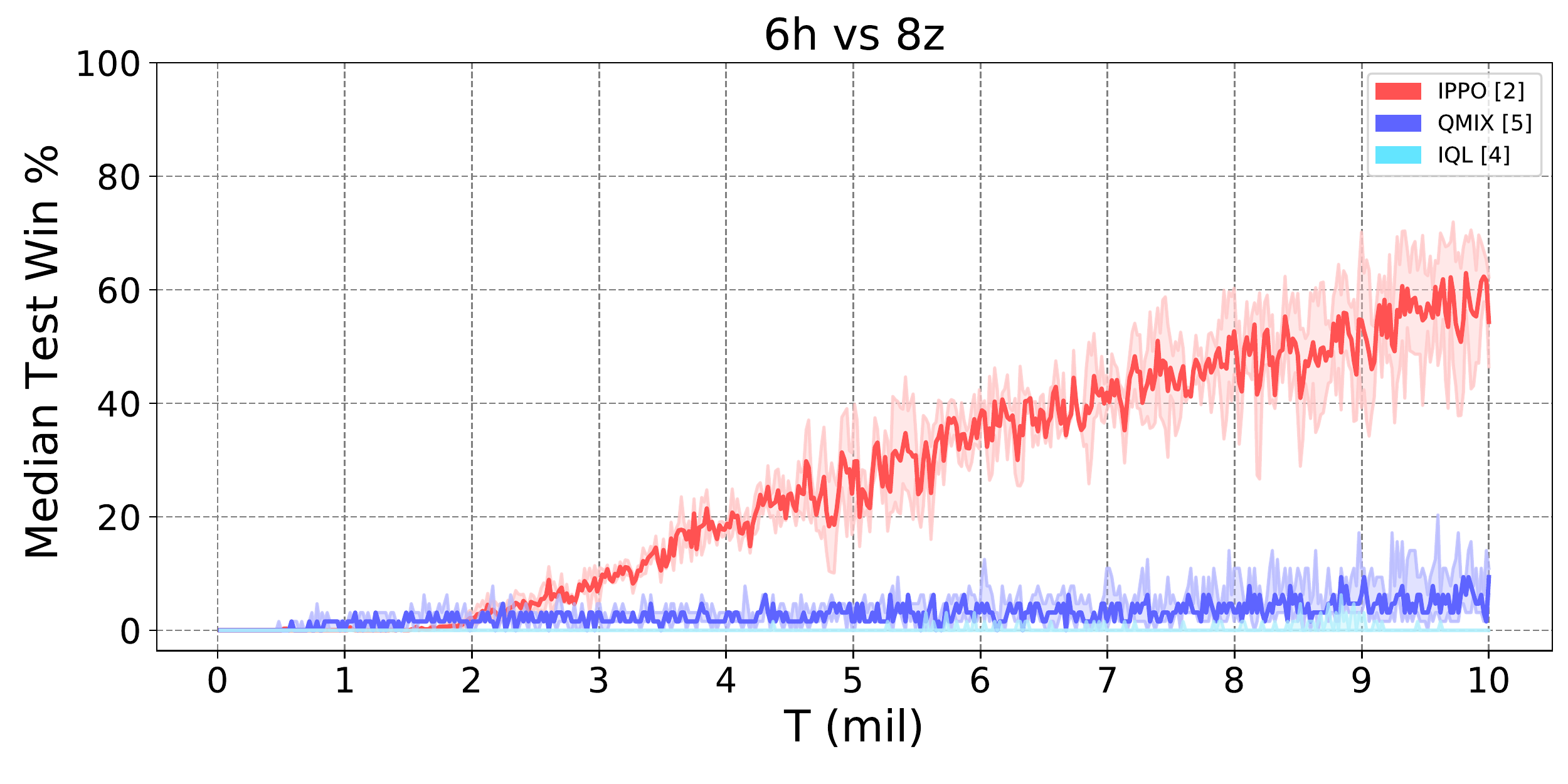} \\ 
(a) corridor & (b) 6h vs 8z \\[6pt] 
\includegraphics[width=65mm]{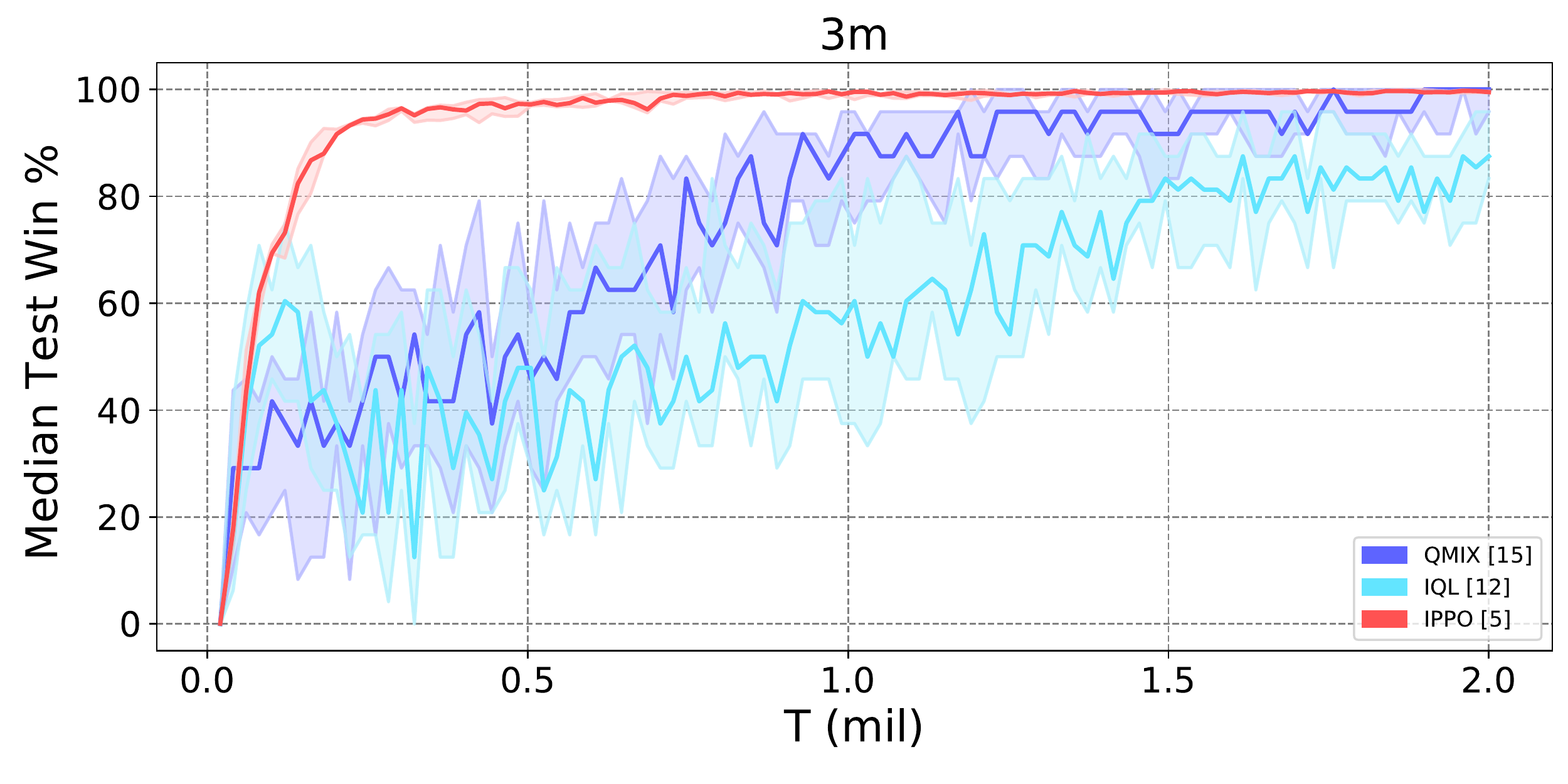} & \includegraphics[width=65mm]{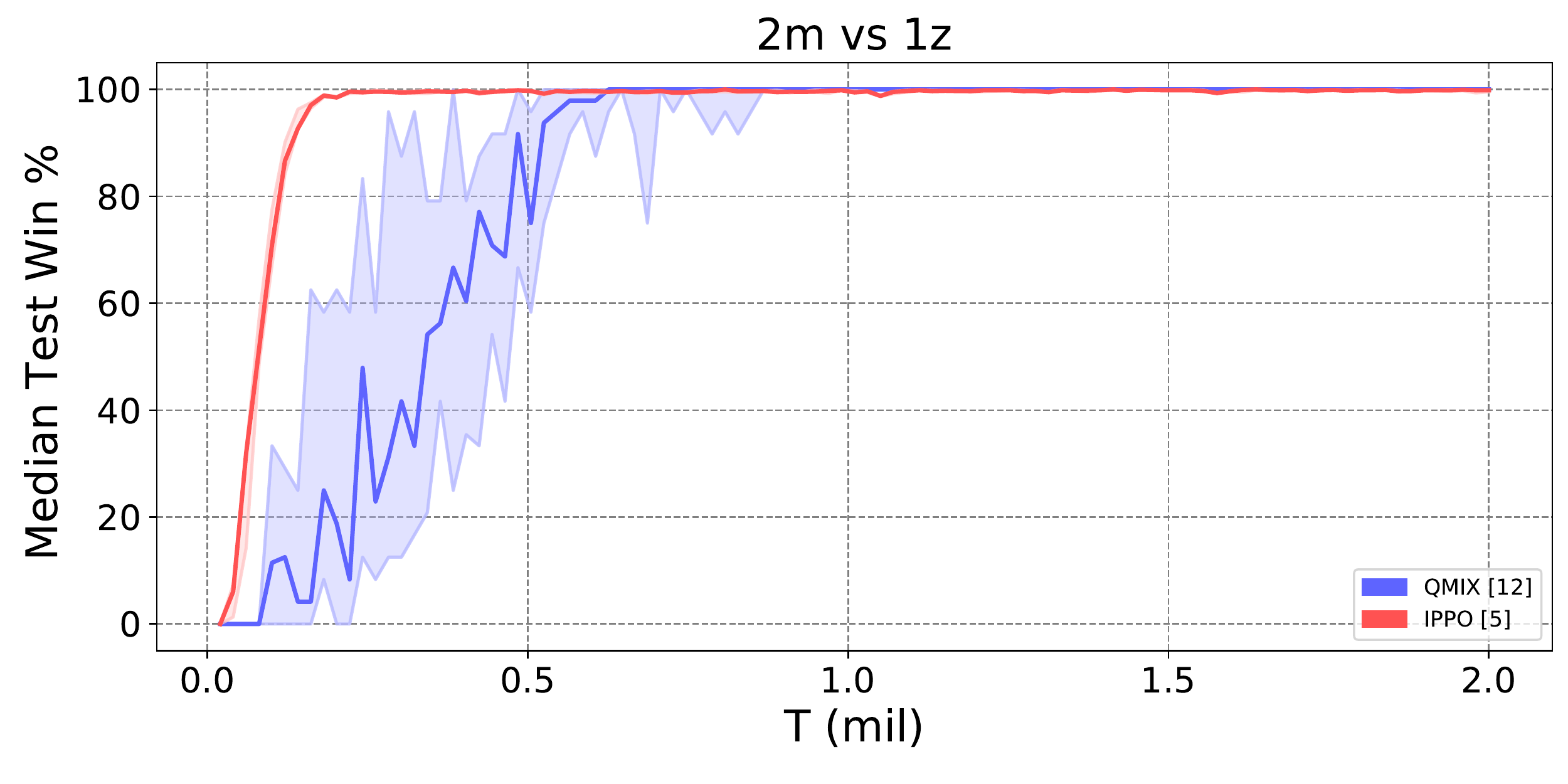}  \\
(c) 3m & (d) 2m vs 1z \\
\multicolumn{2}{c}{\includegraphics[width=65mm]{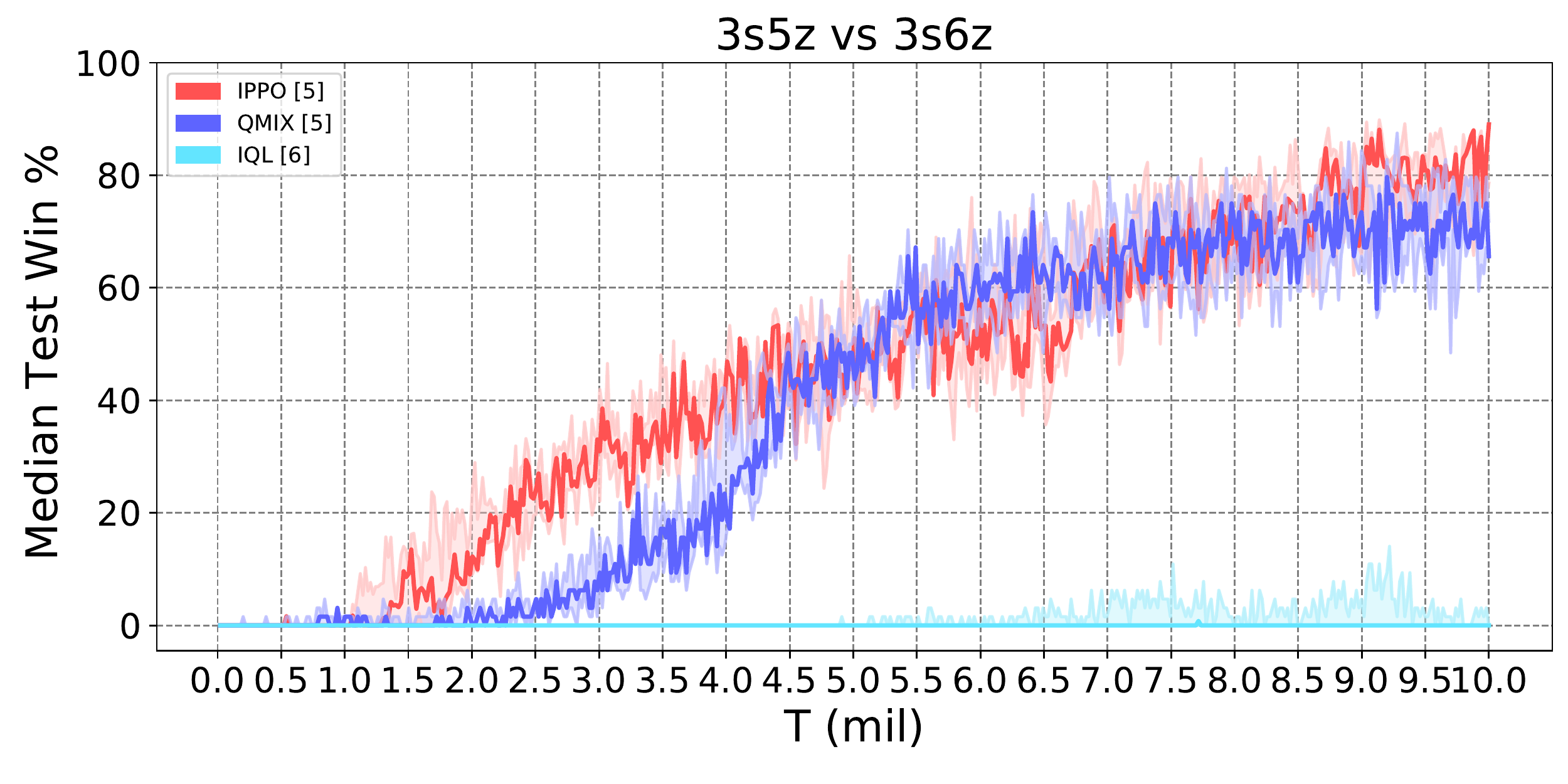}}\\
\multicolumn{2}{c}{(e)}
\end{tabular}
\caption{Results on select SMAC maps, comparing IPPO to QMIX and IQL. Uncertainty regions depict $[0.25,0.75]$ confidence intervals of the median test win rate.}
\end{figure}

\begin{figure}
\label{fig:ippo_tab}
\centering
\begin{tabular}{cccccc}
\toprule
Map     & corridor  & $6h\_vs\_8z$  & $3s5z\_vs\_3s6z$  & $3m$    & $2m\_vs\_1z$ \\   \midrule
IPPO    &  80       & 60            &  90               & 100     & 100 \\ \midrule
MAPPO   & 0         & 8.95          & 80                & 99.875  & 100 \\ \bottomrule
\end{tabular}
\caption{Median win test rates on select SMAC maps at 10M steps, comparing IPPO to MAPPO results reported by \cite{anonymous_benchmarking_2020}.}
\end{figure}


In this section, we evaluate Independent PPO (IPPO) on 16 maps from the StarCraft Multi-Agent Challenge \cite[SMAC]{samvelyan_starcraft_2019} maps, performing only mild hyperparameter tuning for individual maps (see Table \ref{table:paramtable}). SMAC consists of a diverse set of StarCraft II \cite[SCII]{vinyals_starcraft_2017} unit micromanagement tasks of varying difficulty, where a collaborative team of SCII units needs to defeat an enemy team of units controlled by the built-in AI. SCII units consist of a balanced set of both melee and long-range attack units (plus healing units, \textit{medivacs}) and winning strategies often entail precisely coordinated unit movements strategies, including \textit{kiting} \cite{rashid_monotonic_2020}, in order to gain positional advantages. For all our experiments, we use game version $4.6$ and select the hardest AI difficulty level.


\subsection{IPPO Performance}

We benchmark IPPO and a number of ablations on a number of SMAC maps (see Figure \ref{fig:ippo}). 
Despite its simplicity, we find that IPPO significantly outperforms two
strong MARLbaselines that exploit centralized state during learning, namely MAPPO (see Table \ref{fig:ippo_tab} and QMIX
on three difficult maps, namely \textit{3s5z vs 3s6z}, \textit{corridor} and \textit{6h vs 8z}. On \textit{corridor}, IPPO even exceeds the particularly strong performance of MAVEN \cite{mahajan_maven_2020}. In addition, IPPO outperforms QMIX on easy SMAC maps \textit{2m vs 1z} and \textit{3m} and is competitive on a few more (see Appendix \ref{fig:ippomore}).
IPPO is also able to learn maps renowned for their difficulty, such as \textit{3s5z vs 3s6z} and \textit{6h vs 8z}.
We also find that IPPO performance is superior to that of IQL, the other independent learning algorithm tested \cite{rashid_monotonic_2020}, and it is also generally more stable, across a large variety of maps.
Further results may be found in Appendix \ref{sec:appendix}).

\subsection{Role of Centralised Value Functions}

\begin{figure}
\label{fig:cv}
\begin{tabular}{cc}
  \includegraphics[width=65mm]{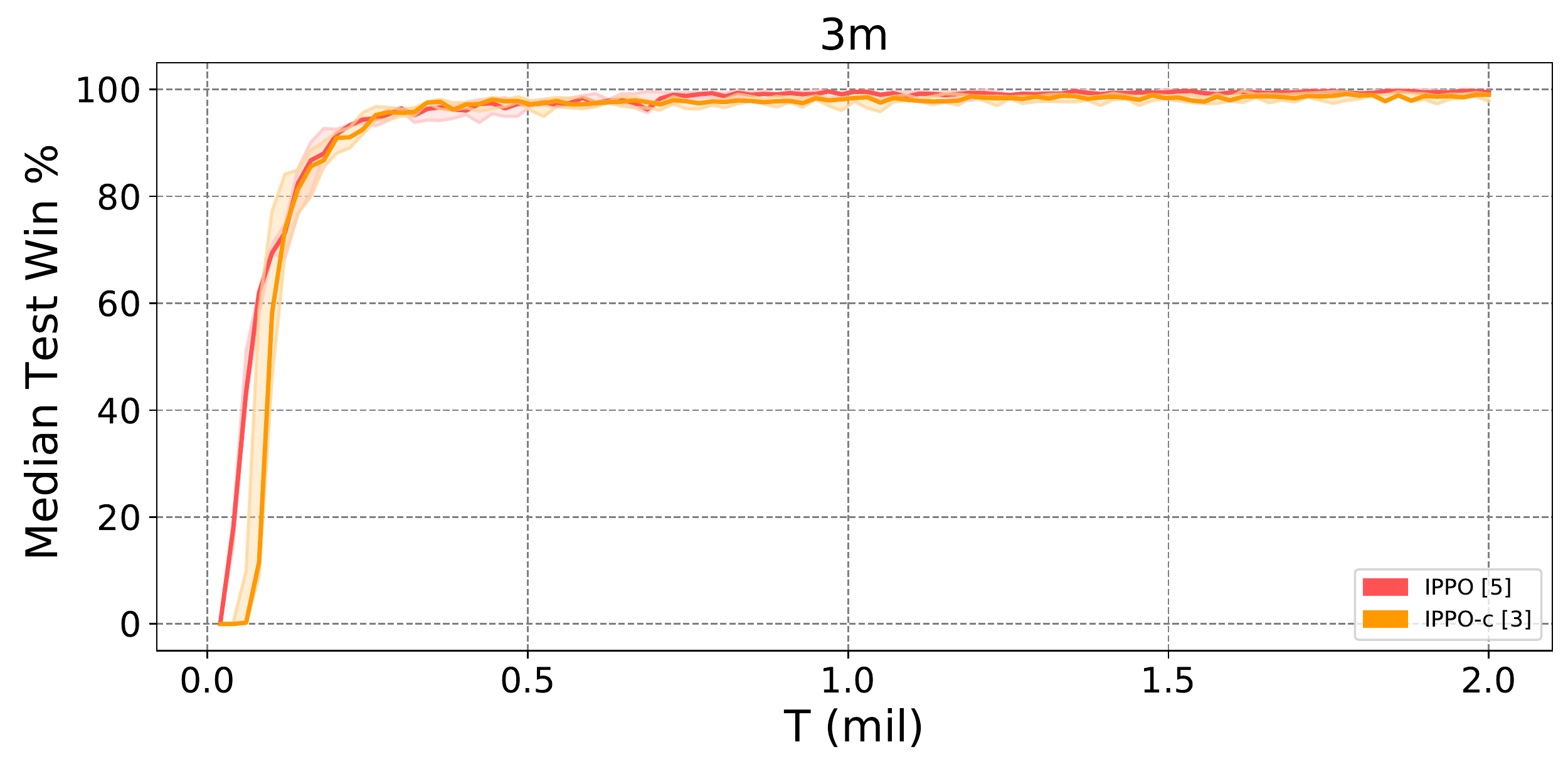} &
  \includegraphics[width=65mm]{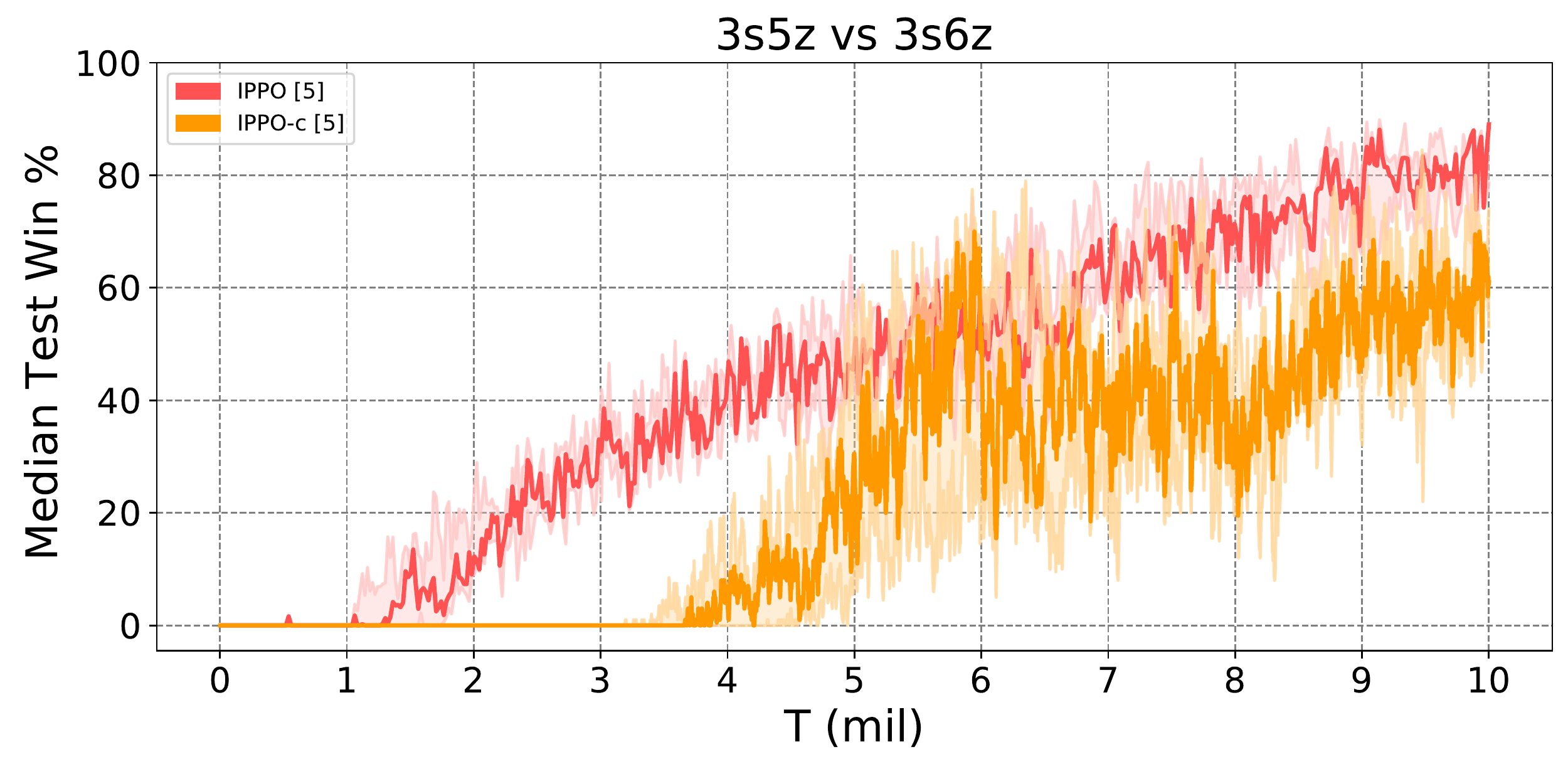}  \\
(a) 3m & (b) 3s5z vs 3s6z \\[6pt]
\includegraphics[width=65mm]{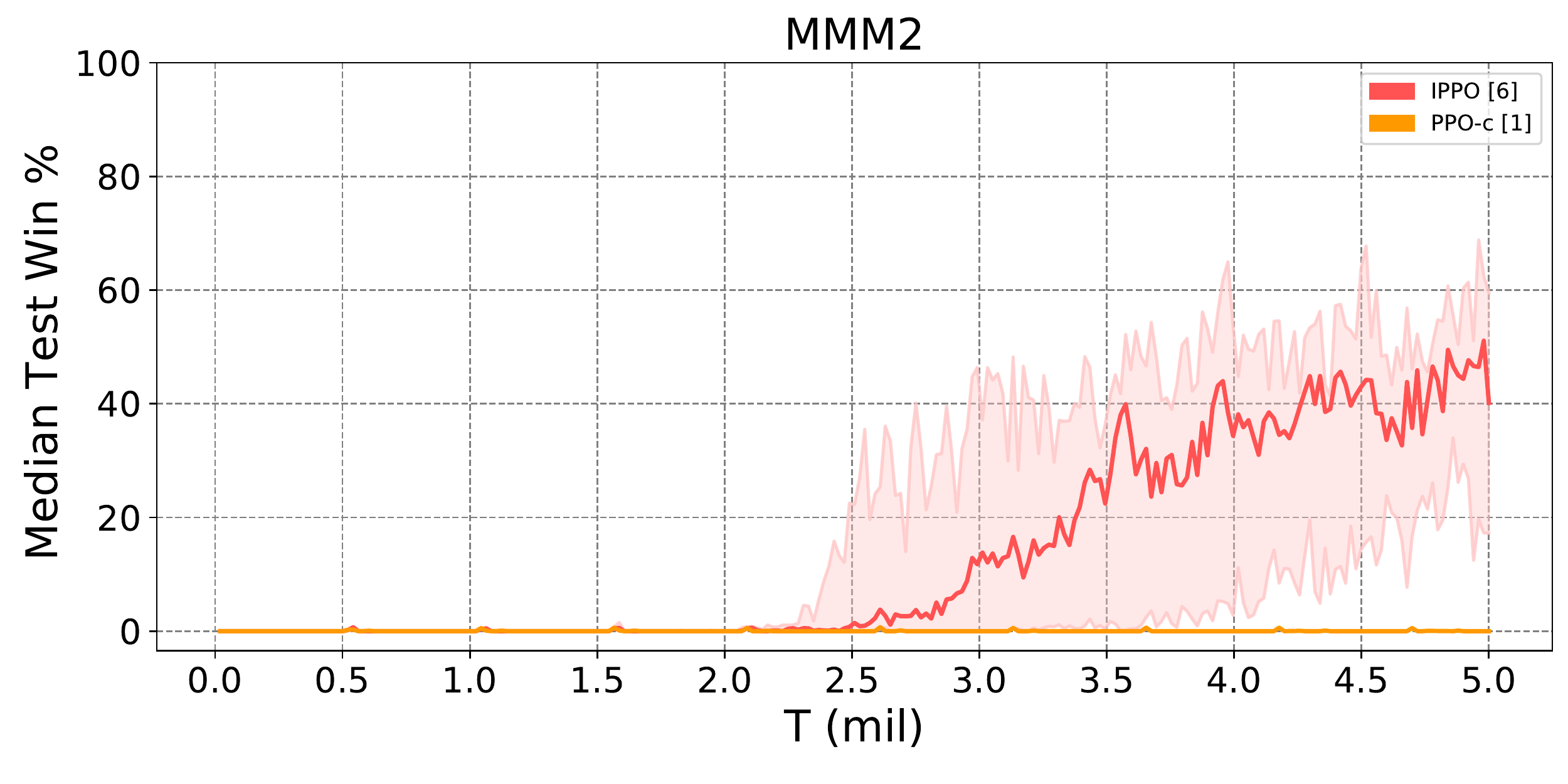}  &
 \includegraphics[width=65mm]{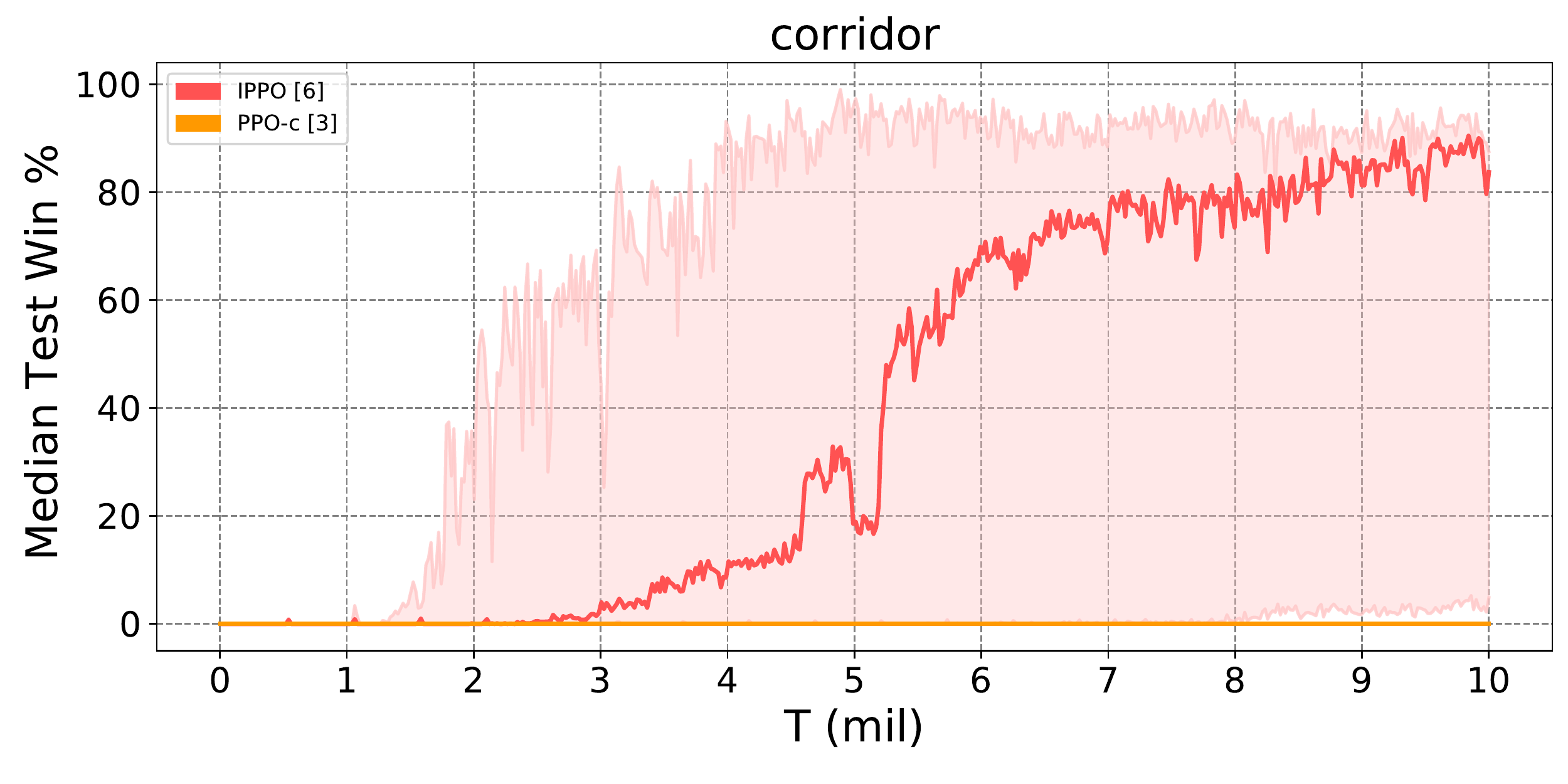} \\
 (c) MMM2 &  (d) corridor \\[6pt]
\end{tabular}
\caption{Results on select SMAC maps comparing IPPO and MAPPO.}
\end{figure}

IPPO exploits centralised training solely through sharing network parameters between agent critics. However, SMAC additionally provides full state information that lifts partial observability of the environment during learning. 
 While at first glance, conditioning the critic on full state information should help simply due to it containing more information, already early empirical research in independent vs centralised learning concludes that, perhaps counter-intuitively, this is not necessarily the case as the larger size of the full state and/or it including information irrelevant to the task may adversely impact learning \cite{tan_multi-agent_1993}. Nevertheless, previous work on SMAC have reported higher performance when the centralised state was being exploited during learning\cite{rashid_monotonic_2020,anonymous_benchmarking_2020}
A direct comparison (see Table \ref{fig:cv}) shows that replacing IPPO's local critics with parameter-sharing critics conditioning on full state information performs substantially worse on a selection of hard SMAC maps, calibrating our implementation with centralised value function performance stated in concurrent work \cite[MAPPO]{anonymous_benchmarking_2020}. 




\subsection{Ablation Studies}
\label{sec:ablation}

\begin{figure}
\label{fig:clip}
\begin{tabular}{cc}
  \includegraphics[width=65mm]{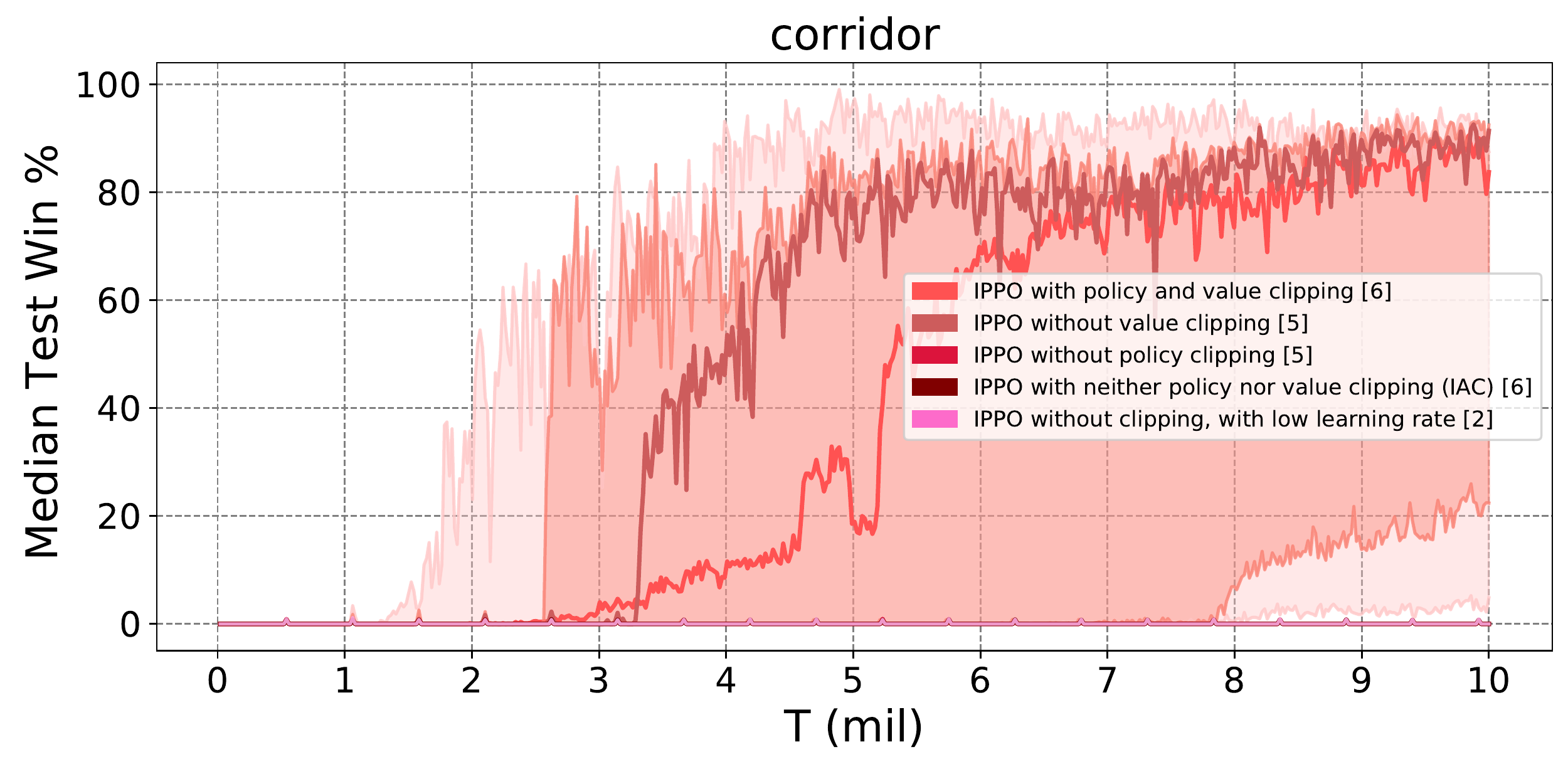} & \includegraphics[width=65mm]{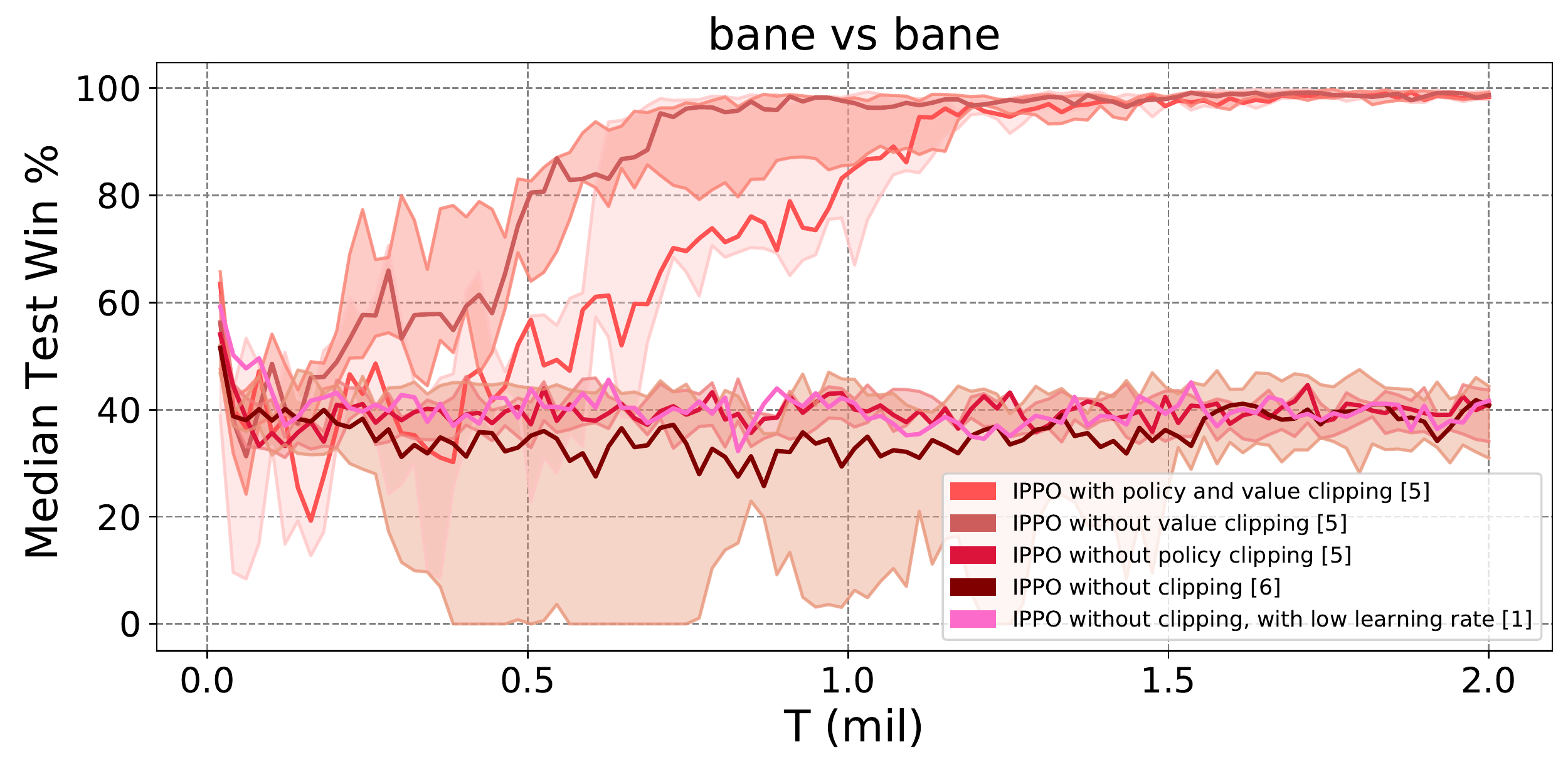}  \\
(a) corridor & (b) bane vs bane  \\[6pt]
\includegraphics[width=65mm]{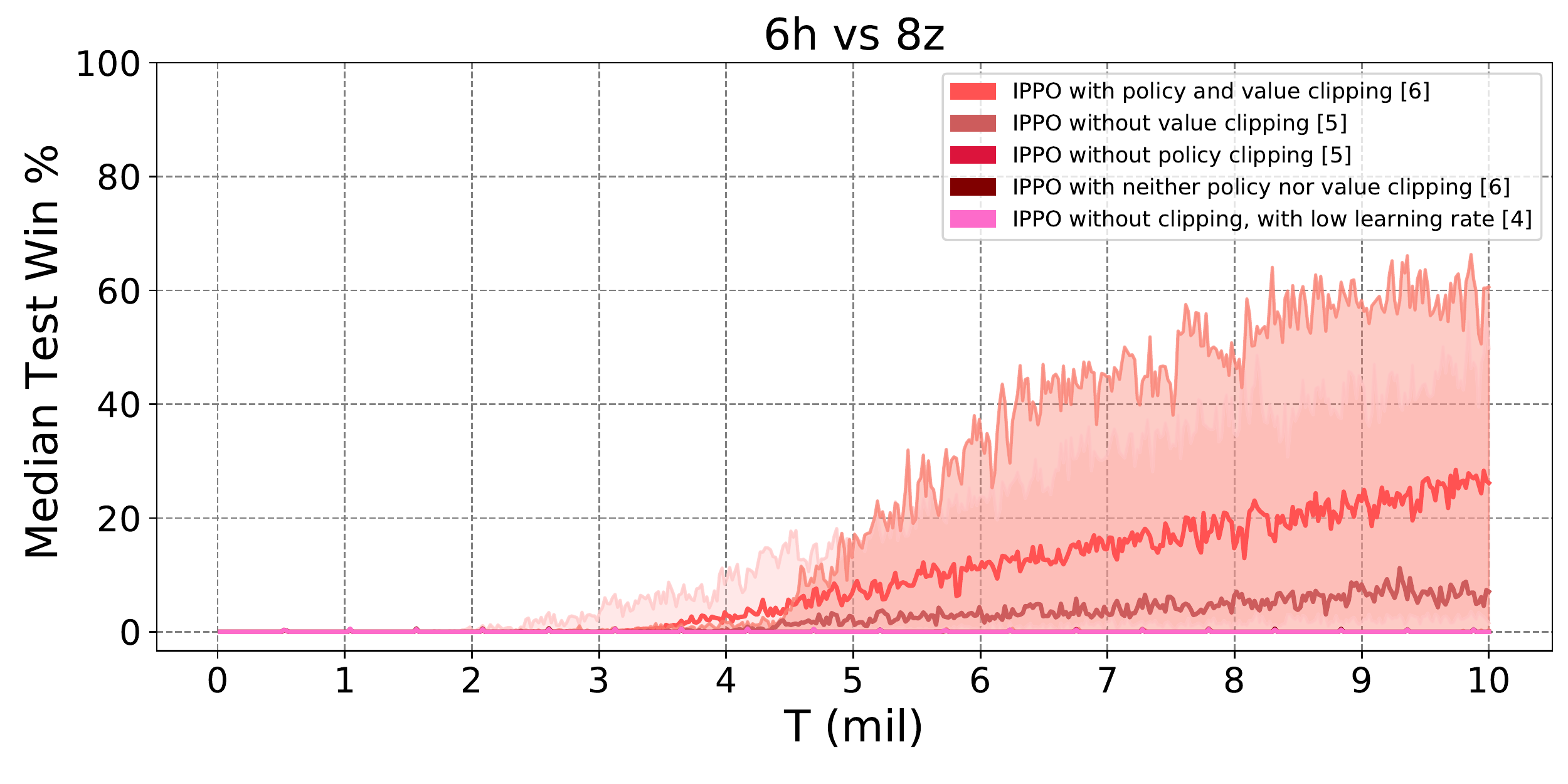}  &
 \includegraphics[width=65mm]{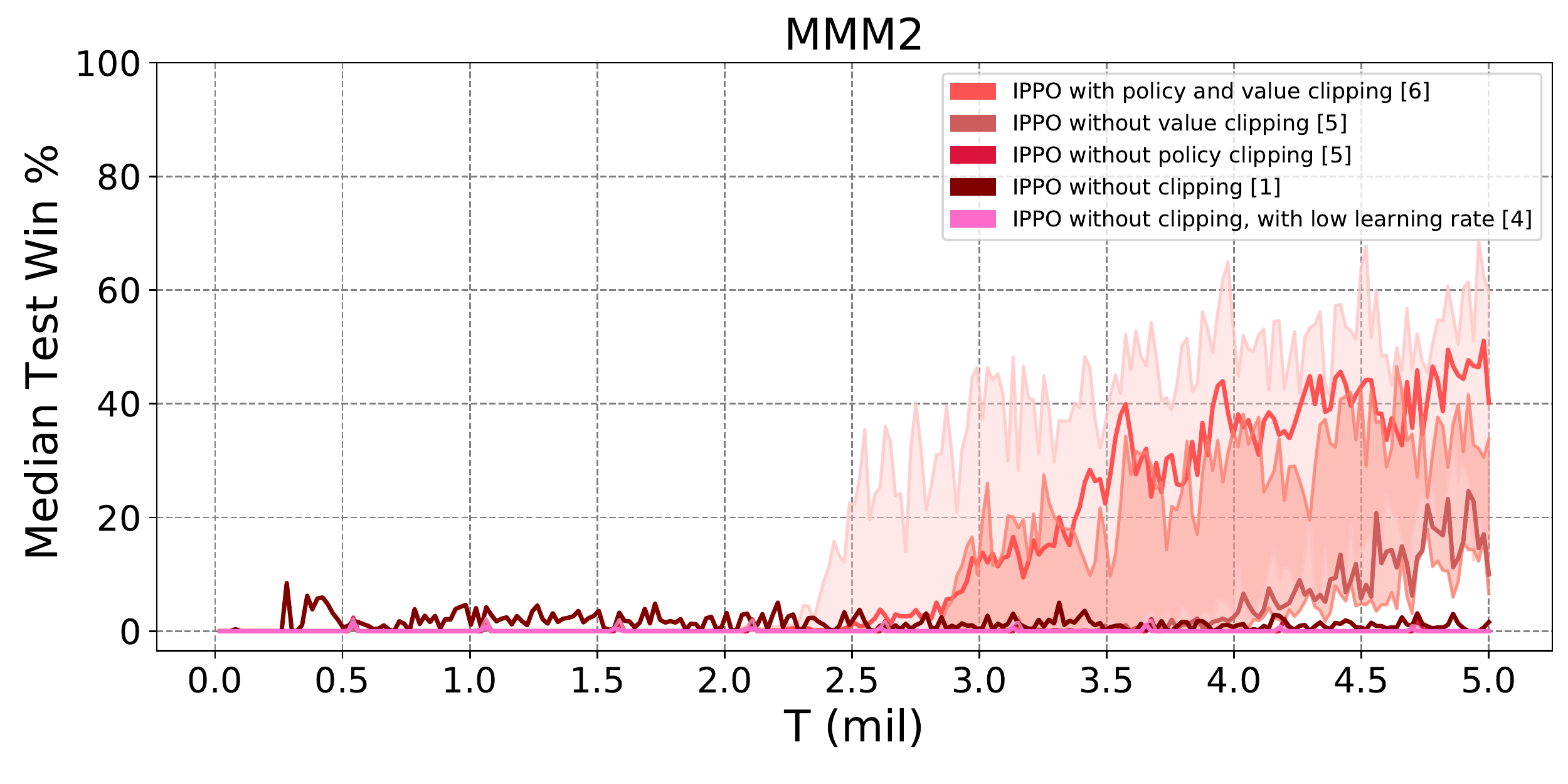} \\
 (c) 6h vs 8z &  (d) MMM2\\
\includegraphics[width=65mm]{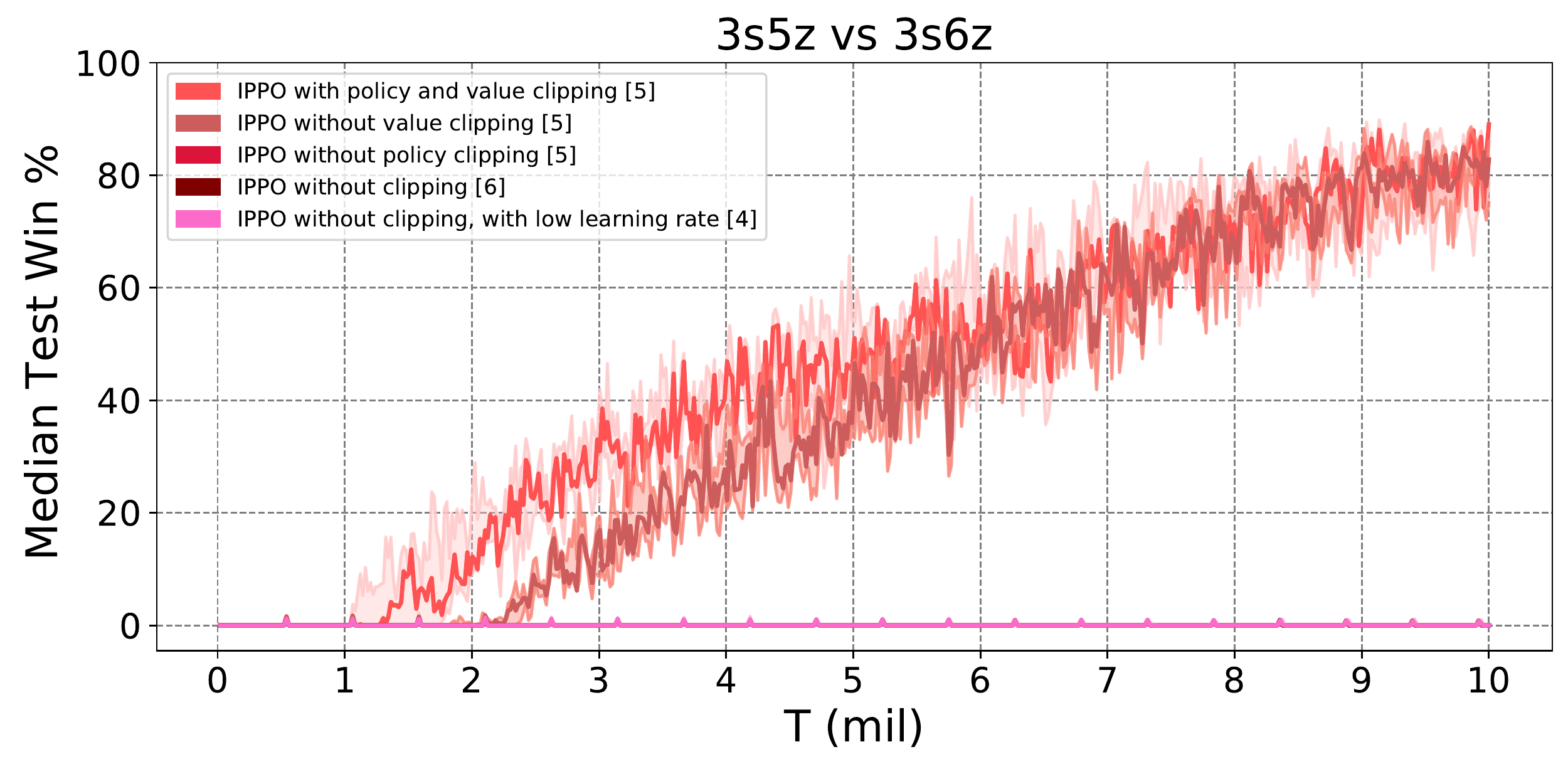}  &
 \includegraphics[width=65mm]{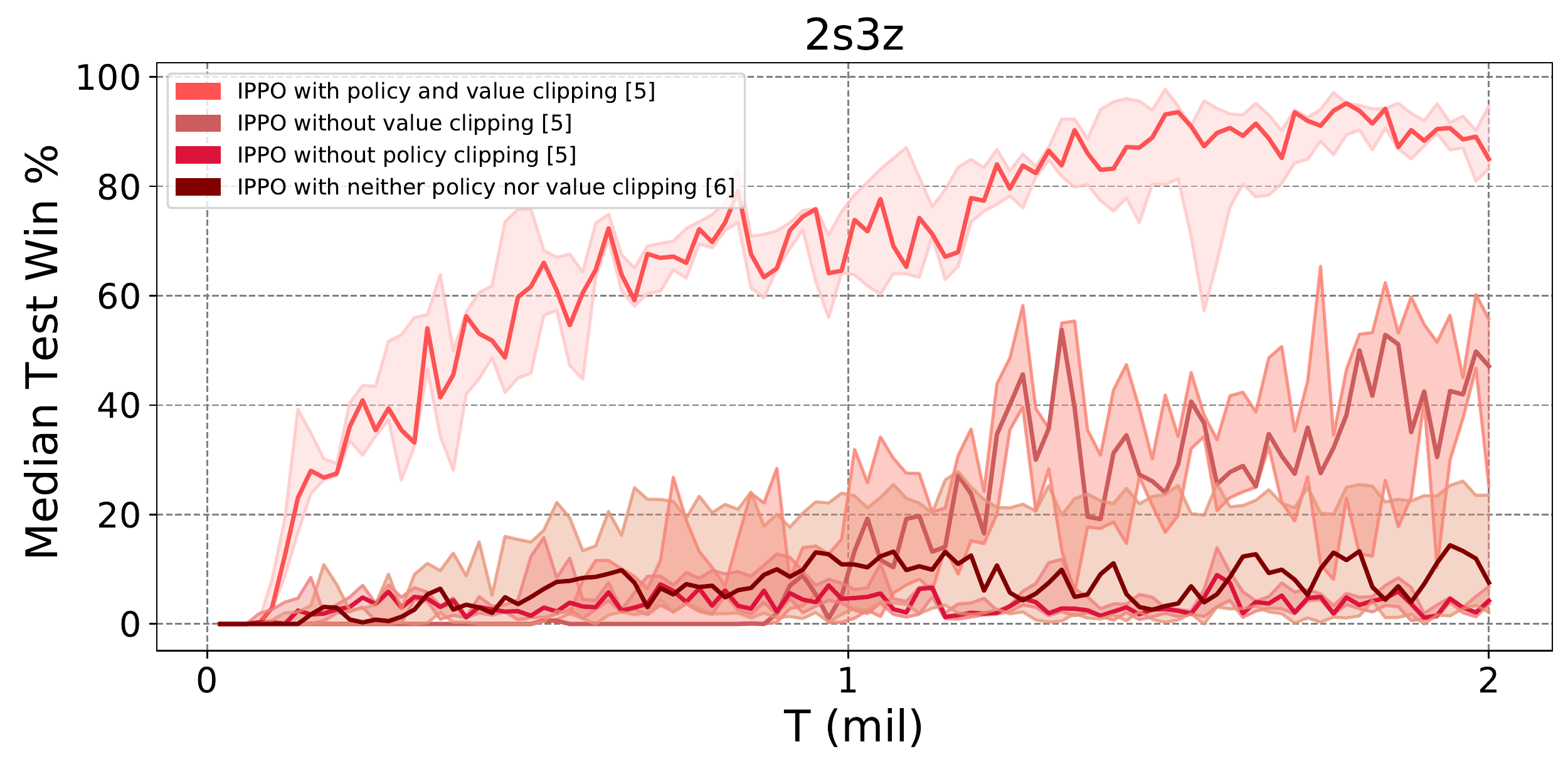} \\
 (e) 3s5z vs 3s6z &  (f) 2s3z\\
\end{tabular}
\caption{Ablation study for IPPO with different combinations of policy and value clipping}
\end{figure}

To understand better why IPPO performs outperforms other independent learning algorithms, including Independent Actor-Critic (IAC) \cite{foerster_counterfactual_2017}, we investigate the performance of a number of ablations (see Figure \ref{fig:clip}). We find that IPPO with neither policy nor value clipping, which corresponds to a variant of IAC \cite{foerster_counterfactual_2017},
 performs poorly across all six SMAC maps studied.  Furthermore, we find that policy clipping is essential to performance. 

In conjunction with policy clipping, we find that value clipping improves performance on some maps (e.g., \textit{corridor} and \textit{MMM2}). The selective usefulness of value function clipping is in line with empirical observations in single-agent settings, which suggest that value function clipping may only be advantageous if the critic estimator suffers from high variance \cite{ikostrikov_value_2018}. 

An alternative explanation for our findings is that IPPO's policy clipping objective reduces the effective policy learning rate, thus stabilising learning. We test this hypothesis by showing reducing the learning rate for IAC.  This does not yield the anticipated performance gains relative to IPPO (see Figure \ref{fig:clip}).

\begin{table}
{\tiny
\begin{center}
  \begin{tabular}{lccccccccccc}
    \toprule
\textbf{map name} & \thead{\tiny \textbf{critic}\\ \tiny\textbf{coef}}& \thead{\tiny \textbf{entropy}\\ \tiny\textbf{coef}} & \textbf{frames} & \textbf{lr} & \thead{\tiny \textbf{mini}\\ \tiny\textbf{epochs}} & \thead{\tiny \textbf{mini}\\ \tiny\textbf{batch}} & \thead{\tiny \textbf{norm}\\ \tiny\textbf{input}} & \textbf{steps num} & \textbf{type} & \textbf{net arch} \\\midrule
10m vs 11m & 2 & 0.005 & 4 & 1e-4 & 4 & 2560 & True & 128 & cnn & [256, 512, 1024]  \\\midrule
1c3s5z & 2 & 0.005 & 4 & 1e-4 & 1 & 4096 & False  &128 & cnn & [64, 128, 256]  \\\midrule
27 vs 30m & 2 & 0.005 & 4 & 1e-4 & 1 & 2560 & False & 64 & cnn & [64, 128, 256]  \\\midrule
2c vs 64zg & 2 & 0.005 & 4 & 1e-4 & 4 & 512 & False & 64 & cnn & [64, 128, 256]  \\\midrule
2m vs 1z & 1 & 0.005 & 1 & 1e-4 & 4 & 1024 & False & 128 & mlp & [256, 128]  \\\midrule
2s3z & 1 & 0.001 & 1 & 5e-4 & 4 & 1536 & True & 128 & cnn & [64, 128, 256]  \\\midrule
2s vs 1sc & 2 & 0.005 & 4 & 1e-4 & 1 & 2048 & False & 128 & cnn & [64, 128, 256]  \\\midrule
3m & 1 & 0.001 & 1 & 5e-4 & 4 & 1536 & True & 128 & mlp & [256, 128]  \\\midrule
3s5z & 2 & 0.005 & 4 & 1e-4 & 1 & 4096 & False & 128 & cnn & [64, 128, 256]  \\\midrule
3s5z vs 3s6z & 2 & 0.005 & 4 & 1e-4 & 1 & 4096 & False & 128 & cnn & [64, 128, 256]  \\\midrule
3s vs 5z & 2 & 0.005 & 4 & 1e-4 & 1 & 1536 & False & 128 & cnn & [64, 128, 256]  \\\midrule
5m vs 6m & 2 & 0.005 & 4 & 1e-4 & 1 & 2560 & False & 128 & cnn & [64, 128, 256]  \\\midrule
6h vs 8z & 2 & 0.005 & 4 & 1e-4 & 1 & 3072 & False & 128 & cnn & [64, 128, 256]  \\\midrule
MMM & 2 & 0.005 & 4 & 1e-4 & 1 & 2560 & False & 64 & cnn & [64, 128, 256]  \\\midrule
MMM2 & 2 & 0.005 & 4 & 1e-4 & 1 & 2560 & False & 64 & cnn & [64, 128, 256]  \\\midrule
bane vs bane & 2 & 0.005 & 4 & 1e-4 & 1 & 3072 & False & 128 & cnn & [64, 128, 256]  \\\midrule
corridor & 2 & 0.005 & 4 & 1e-4 & 1 & 3072 & False & 128 & cnn & [64, 128, 256]  \\
    \bottomrule
  \end{tabular}
\end{center}
}

\caption{IPPO hyperparameters across different SMAC maps. Fixed hyperparameters: $\gamma=0.99$, $\lambda=0.95$, $e_clip=0.2$, $grad \ norm=0.5$ and $\epsilon=0.2$. We do not employ any learning rate schedule, and always normalize the advantage function. We fix the number of actors to $8$. \textit{Net arch} indicates successive channels across layers for \textit{conv1d} architecture, and number of activations for \textit{mlp}. All convolutions have stride $3$.}

\label{table:paramtable}
\end{table}

%% file: sections/06-conclusion.tex
\section{Discussion}

Our empirical results give rise to a number of interesting insights 
about the utility of IPPO in relation to other, state-of-the-art MARL algorithms, in particular those relying on value factorization methods, as well as the role of SMAC in future MARL research. 

First, IPPO outperforms a number of algorithms using centralised state, including QMIX, MAPPO,
and MAVEN, as well as both independent learning algorithms IAC and IQL, on a number of both hard and easy SMAC maps. 
This is surprising, given the community's recent focus on developing MARL algorithms that can exploit state during centralised training.

Second, PPO's optimisation objective, in particular policy clipping, is  crucial to performance in cooperative deep multi-agent reinforcement learning on popular benchmark environment SMAC. 
According to our empirical ablation studies, this effect cannot be explained through a decrease in the effective learning rate alone. Given IPPO's improved learning stability over IAC and IQL, it seems that its (approximate) surrogate objective might mitigate certain forms of environment non-stationarity that other independent learning algorithms are prone to, e.g., by suppressing updates catastrophic to performance.

Third, we show that, at least on some hard SMAC maps, the currently best performance is achieved by an algorithm that does not require the learning of neither joint nor centralised value functions, and does not exploit central state information in any other way. 
This implies that the fundamental obstacles to independent learning methods posed by certain pathological matrix games \cite{claus_dynamics_1998} are not present in such SMAC tasks. A possible explanation is that the sequential nature of SMAC tasks allows the decomposition of tricky simultaneous coordination tasks into a temporal sequence of easier coordination tasks.
The fact that the common learning pathology of relative overgeneralisation  does not seem to be present in these SMAC tasks (see Fig.\ \ref{fig:over} in the Appendix for an empirical evidence that IPPO -- like many value factorisation algorithms -- is in principle prone to relative overgeneralisation). 


Finally, it remains unclear what exactly the value of central state information is in SMAC. Empirical ablation studies in value factorisation algorithms, such as QMIX, clearly show that using central state information on top of local observations can accelerate training. However, IPPO's strong performance on particularly hard SMAC maps and, in particular, its outperformance of MAPPO on these, raise the question as to what exactly makes central state information useful in QMIX.


\section{Conclusion}

In this paper, we studed the empirical performance of Independent PPO (IPPO), a multi-agent variant of the popular PPO algorithm. Despite the purported limitations of independent learning for cooperative MARL tasks, we found that IPPO performs competitively on a range of state-of-the-art benchmark tasks, outperforming state-of-the-art value factorization methods on some maps. A number of ablation studies indicated that PPO's policy clipping objective is crucial to performance, and that the value of central state information in SMAC is unclear. Furthermore, our results raise the question of whether relative overgeneralization really matters in practice: Despite of the diversity of its scenarios, it does not seem to be an obstacle to independent learning in SMAC.  

As a consequence, we suggest that the MARL community revisit the question of whether relative overgeneralisation really matters in practice. 
Secondly, these results suggest it would be fruitful to further improve independent learning approaches such as IPPO, rather than solely focusing research on joint value function factorisation.